\pdfoutput=1
\documentclass[10pt,logo,copyright]{techreport}
\usepackage[authoryear,sort&compress,round]{natbib}

\usepackage[utf8]{inputenc}
\usepackage[T1]{fontenc}
\usepackage{fancyhdr}
\usepackage{ifthen}
\usepackage{booktabs}

\usepackage{pifont} 
\usepackage[most]{tcolorbox}
\usepackage{enumitem}
\usepackage[table]{xcolor}
\usepackage{subcaption}
\usepackage{amssymb}
\usepackage{array}
\usepackage{longtable}

\usepackage{listings}
\usepackage{xcolor}
\usepackage{textcomp}

\usepackage{parskip}
\usepackage{url}
\usepackage{hyperref}
\usepackage{booktabs}
\usepackage{amsfonts}
\usepackage{nicefrac}
\usepackage{microtype}
\usepackage{xcolor}
\usepackage[dvipsnames]{xcolor}
\usepackage{graphicx}
\usepackage{tabularx}
\usepackage{makecell}
\usepackage{float}
\usepackage[section]{placeins}
\usepackage{wrapfig}

\usepackage{amsmath,amsfonts,bm,bbm}
\usepackage{multirow}

\definecolor{best_green}{HTML}{DDF4DD}
\definecolor{second_purple}{HTML}{E4E0FF}

\newcommand{\best}[1]{\cellcolor{best_green}\textbf{#1}}
\newcommand{\second}[1]{\cellcolor{second_purple}\textbf{#1}}
\newcommand{\refcell}[1]{\textcolor{gray!60}{#1}}
\DeclareRobustCommand{\besttext}[1]{%
  {\setlength{\fboxsep}{0.3pt}%
  \colorbox{best_green}{\strut #1}}%
}

\DeclareRobustCommand{\secondtext}[1]{%
  {\setlength{\fboxsep}{0.3pt}%
  \colorbox{second_purple}{\strut #1}}%
}
\definecolor{MyRed}{RGB}{216,56,58}
\definecolor{MyBlue}{RGB}{69,127,228}
\definecolor{MyOrange}{RGB}{238,130,47}
\definecolor{MyGreen}{RGB}{90,194,149}
\definecolor{lightblue}{rgb}{0.46,0.73,0.00}

\definecolor{jsonbg}{HTML}{F7F7F7}
\definecolor{jsonframe}{HTML}{D9D9D9}
\definecolor{jsontext}{HTML}{24292F}
\definecolor{jsonstring}{HTML}{0A3069}
\definecolor{jsonkeyword}{HTML}{8250DF}

\lstdefinelanguage{myjson}{
  morestring=[b]",
  stringstyle=\color{jsonstring},
  morekeywords={true,false,null},
  keywordstyle=\color{jsonkeyword},
  sensitive=false
}

\lstdefinestyle{jsonstyle}{
  language=myjson,
  backgroundcolor=\color{jsonbg},
  basicstyle=\ttfamily\scriptsize\color{jsontext},
  numbers=none,
  frame=single,
  framerule=0.4pt,
  rulecolor=\color{jsonframe},
  framesep=6pt,
  breaklines=true,
  breakatwhitespace=false,
  showstringspaces=false,
  columns=fullflexible,
  keepspaces=true,
  upquote=true,
  captionpos=b,
  aboveskip=0.8em,
  belowskip=0.8em,
  xleftmargin=0.3em,
  xrightmargin=0.3em
}

\usepackage[most]{tcolorbox}
\tcbuselibrary{listings,breakable,skins}
\usepackage{textcomp}

\definecolor{promptbg}{HTML}{F7F7F7}
\definecolor{promptframe}{HTML}{D9D9D9}
\definecolor{prompttitle}{RGB}{105,105,105}
\definecolor{prompttext}{HTML}{24292F}

\lstdefinestyle{promptstyle}{
  basicstyle=\ttfamily\scriptsize\color{prompttext},
  numbers=none,
  breaklines=true,
  breakatwhitespace=false,
  showstringspaces=false,
  columns=fullflexible,
  keepspaces=true,
  upquote=true
}

\newtcblisting[auto counter, number within=section]{promptbox}[2][]{
  enhanced,
  breakable,
  listing only,
  title={Prompt Box~\thetcbcounter: #2},
  label={#1},
  colback=promptbg,
  colframe=promptframe,
  colbacktitle=prompttitle,
  coltitle=white,
  fonttitle=\scriptsize\bfseries,
  arc=0.8mm,
  boxrule=0.4pt,
  boxsep=1mm,
  left=1mm,
  right=1mm,
  top=1mm,
  bottom=1mm,
  listing options={style=promptstyle}
}
\PassOptionsToPackage{authoryear,round}{natbib}

\definecolor{figcitecolor}{HTML}{30b776}
\usepackage{hyperref}
\hypersetup{
    colorlinks=true,
    linkcolor=figcitecolor,
    citecolor=tongyipurple,
    urlcolor=tongyicite
}
\setcitestyle{authoryear,round,semicolon}
\bibpunct{(}{)}{;}{a}{,}{,}

\newtcolorbox{questionbox}{
  colback=blue!3,
  colframe=blue!25,
  boxrule=0.4pt,
  arc=2pt,
  left=4pt,
  right=4pt,
  top=3pt,
  bottom=3pt,
  before skip=4pt,
  after skip=4pt
}

\usepackage{tabularx}
\usepackage{array}
\usepackage{multirow}
\usepackage{graphicx}
\usepackage[table]{xcolor}

\definecolor{improvegreen}{HTML}{557E76}

\definecolor{stdmbg}{RGB}{249,249,252}
\definecolor{stdmtitle}{RGB}{236,232,248}
\definecolor{stdmframe}{RGB}{218,218,228}

\definecolor{titlebg}{RGB}{242,242,246}
\definecolor{titlefg}{RGB}{35,35,35}
\definecolor{tagbg}{RGB}{245,245,245}
\definecolor{tagfg}{RGB}{45,45,45}

\definecolor{topkey}{RGB} {38,84,124}
\definecolor{midkey}{RGB}{112,72,190}
\definecolor{lowkey}{RGB}{0,128,128}
\definecolor{valc}{RGB}{184,96,24}

\definecolor{commentgreen}{RGB}{24,102,48}
\definecolor{tagbg}{RGB}{235,246,239}

\usepackage{wrapfig}
\newtcolorbox{stdmbox}[1][]{
  enhanced,
  colback=stdmbg,
  colframe=stdmframe,
  colbacktitle=stdmtitle,
  coltitle=black,
  fonttitle=\footnotesize,
  title=Base STDM Structure for Our Executable Memory,
  arc=1.0pt,
  boxrule=0.35pt,
  left=1.5pt,
  right=0.0pt,
  top=1.5pt,
  bottom=1.5pt,
  boxsep=1.5pt,
  titlerule=0pt,
  toptitle=1.5pt,
  bottomtitle=1.5pt,
  lefttitle=3pt,
  righttitle=8pt,
  #1
}

\definecolor{fieldbg}{RGB}{238,245,255}
\definecolor{fieldfg}{RGB}{38,84,124}

\definecolor{schemabg}{RGB}{245,240,255}
\definecolor{schemafg}{RGB}{112,72,190}

\definecolor{valuebg}{RGB}{255,246,232}
\definecolor{valuefg}{RGB}{184,96,24}

\definecolor{statusbg}{RGB}{235,246,239}
\definecolor{statusfg}{RGB}{24,102,48}

\definecolor{memtagbg}{RGB}{238,238,238}
\definecolor{memtagfg}{RGB}{30,30,30}

\newtcbox{\fieldtag}{
  on line,
  colback=fieldbg,
  colframe=fieldbg,
  boxrule=0pt,
  arc=1pt,
  left=1pt,
  right=1pt,
  top=0.5pt,
  bottom=0.5pt,
  boxsep=0pt,
  fontupper=\ttfamily\bfseries\footnotesize\color{fieldfg}
}

\newtcbox{\schematag}{
  on line,
  colback=schemabg,
  colframe=schemabg,
  boxrule=0pt,
  arc=1pt,
  left=1pt,
  right=1pt,
  top=0.5pt,
  bottom=0.5pt,
  boxsep=0pt,
  fontupper=\ttfamily\bfseries\footnotesize\color{schemafg}
}

\newtcbox{\valuetag}{
  on line,
  colback=valuebg,
  colframe=valuebg,
  boxrule=0pt,
  arc=1pt,
  left=1pt,
  right=1pt,
  top=0.5pt,
  bottom=0.5pt,
  boxsep=0pt,
  fontupper=\ttfamily\footnotesize\color{valuefg}
}

\newtcbox{\statustag}{
  on line,
  colback=statusbg,
  colframe=statusbg,
  boxrule=0pt,
  arc=1pt,
  left=1pt,
  right=1pt,
  top=0.5pt,
  bottom=0.5pt,
  boxsep=0pt,
  fontupper=\ttfamily\bfseries\footnotesize\color{statusfg}
}
\definecolor{question}{RGB}{245,245,245} 

\definecolor{lightpurple}{RGB}{245,240,255}

\definecolor{appCalendarBg}{RGB}{232,240,254}
\definecolor{appCalendarFg}{RGB}{52,87,178}

\definecolor{appRetroBg}{RGB}{243,233,255}
\definecolor{appRetroFg}{RGB}{126,87,194}

\definecolor{appClockBg}{RGB}{255,243,224}
\definecolor{appClockFg}{RGB}{191,110,0}

\definecolor{appContactsBg}{RGB}{232,245,233}
\definecolor{appContactsFg}{RGB}{46,125,50}

\definecolor{appMarkorBg}{RGB}{240,240,240}
\definecolor{appMarkorFg}{RGB}{90,90,90}

\definecolor{appPhoneBg}{RGB}{227,242,253}
\definecolor{appPhoneFg}{RGB}{2,119,189}

\definecolor{appTasksBg}{RGB}{255,235,238}
\definecolor{appTasksFg}{RGB}{173,20,87}

\definecolor{appChromeBg}{RGB}{255,248,225}
\definecolor{appChromeFg}{RGB}{180,120,0}

\definecolor{appMessagesBg}{RGB}{232,245,233}
\definecolor{appMessagesFg}{RGB}{56,142,60}

\definecolor{appExpenseBg}{RGB}{252,228,236}
\definecolor{appExpenseFg}{RGB}{194,24,91}

\definecolor{appFilesBg}{RGB}{232,244,253}
\definecolor{appFilesFg}{RGB}{25,118,210}

\definecolor{appBroccoliBg}{RGB}{241,248,233}
\definecolor{appBroccoliFg}{RGB}{104,159,56}

\definecolor{appGalleryBg}{RGB}{243,229,245}
\definecolor{appGalleryFg}{RGB}{123,31,162}

\definecolor{appVlcBg}{RGB}{255,243,224}
\definecolor{appVlcFg}{RGB}{245,124,0}

\newtcbox{\applabelbox}[1][]{
  on line,
  boxrule=0pt,
  arc=4pt,
  left=4pt,
  right=4pt,
  top=1.5pt,
  bottom=1.5pt,
  boxsep=0pt,
  colframe=white,
  #1
}

\newcommand{\appCalendar}{\applabelbox[colback=appCalendarBg]{\textcolor{appCalendarFg}{Calendar}}}
\newcommand{\appRetroMusic}{\applabelbox[colback=appRetroBg]{\textcolor{appRetroFg}{Music}}}
\newcommand{\appClock}{\applabelbox[colback=appClockBg]{\textcolor{appClockFg}{Clock}}}
\newcommand{\appContacts}{\applabelbox[colback=appContactsBg]{\textcolor{appContactsFg}{Contacts}}}
\newcommand{\appMarkor}{\applabelbox[colback=appMarkorBg]{\textcolor{appMarkorFg}{Markor}}}
\newcommand{\appPhone}{\applabelbox[colback=appPhoneBg]{\textcolor{appPhoneFg}{Phone}}}
\newcommand{\appTasks}{\applabelbox[colback=appTasksBg]{\textcolor{appTasksFg}{Tasks}}}
\newcommand{\appChrome}{\applabelbox[colback=appChromeBg]{\textcolor{appChromeFg}{Chrome}}}
\newcommand{\appMessages}{\applabelbox[colback=appMessagesBg]{\textcolor{appMessagesFg}{Messages}}}
\newcommand{\appExpense}{\applabelbox[colback=appExpenseBg]{\textcolor{appExpenseFg}{Expense Book}}}
\newcommand{\appFiles}{\applabelbox[colback=appFilesBg]{\textcolor{appFilesFg}{Files}}}
\newcommand{\appBroccoli}{\applabelbox[colback=appBroccoliBg]{\textcolor{appBroccoliFg}{Broccoli}}}
\newcommand{\appGallery}{\applabelbox[colback=appGalleryBg]{\textcolor{appGalleryFg}{Gallery}}}
\newcommand{\appVLC}{\applabelbox[colback=appVlcBg]{\textcolor{appVlcFg}{VLC}}}

\title{What Memory Do GUI Agents Really Need? From Passive Records to Active Task-Driving States}
\vspace{-1.0em}
 \author{Chen Liu$^{2*}$, ~~Ling Chen$^{2}$, ~~Hanzhang Zhou$^{1}$$^\dagger$, ~~Xu Zhang$^{1}$, ~~Quyu Kong$^{1}$, ~~Panrong Tong$^{1}$, ~~Wenhao Wang$^{1}$, ~~Xin Yu$^{3}$, ~~Steven Hoi$^{1}$, ~~Yue Wang$^{1}$$^\dagger$}

\begin{document}

\maketitle


\begin{figure*}[htp]
\centering
\vspace{-2.5em}
\includegraphics[width=0.98\textwidth]{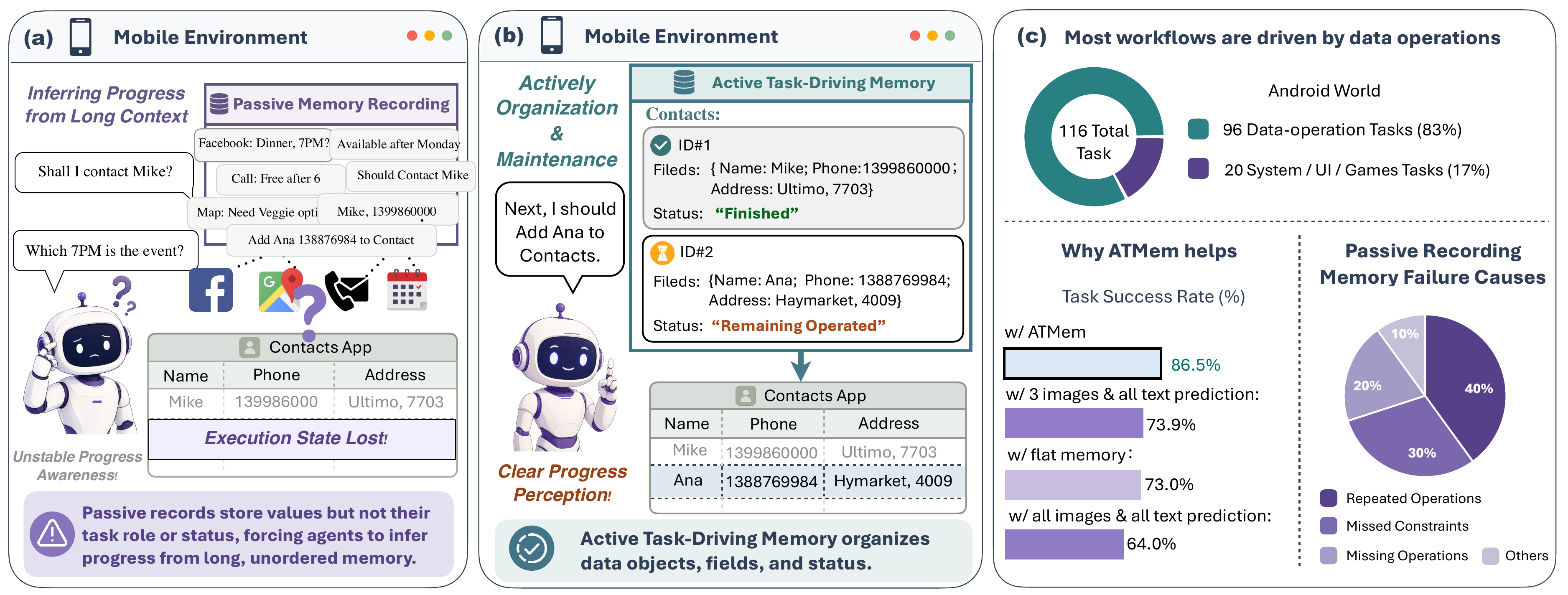}
\vspace{-0.5em}
\caption{\textbf{Motivation.}
Passive records preserve past snippets but do not provide stable execution-state awareness, leading to missed, repeated, or over-scoped operations.
(c) AndroidWorld statistics show that 83\% of tasks involve data operations.
(d) With the same GPT-5 planner and UIIns grounding framework, ATMem improves over flat-memory and full-history baselines, and failure analysis attributes flat-memory errors mainly to repeated operations, missed operations, and forgotten constraints.}
\label{fig:teaser}
\end{figure*}

\begin{abstract}

Mobile GUI agents increasingly face long-horizon tasks that require reading, updating, and reusing task-relevant data across pages and applications. 
Existing methods treat memory largely as passive storage, where past observations are accumulated and retrieved when needed. 
Yet retrieving a value does not reveal its current role in the workflow. 
The agent must still infer from accumulated records whether the value should be used now, has already been used, or must wait for a later dependency. 
This implicit reconstruction becomes unreliable in long trajectories with repeated values, distractors, and outdated states, causing repeated or missed operations. 
To address this, we propose Active Task Driving Memory (ATMem), which shifts GUI-agent memory from passive storage to an actively maintained execution state. 
ATMem maintains task-relevant information as a continually updated execution state that links each value to its role and current status, enabling action selection based on the current workflow state.
While supervised fine-tuning enables the agent to construct ATMem, it does not teach when ATMem is beneficial. 
We therefore introduce \textbf{STR-GRPO}, an online reinforcement learning method that encourages selective use of ATMem based on its contribution to task completion.
STR-GRPO contrasts memory-on and memory-off rollouts to estimate when memory use improves execution, while memory-cost-aware reward discourages costly memory usage that does not improve execution.
To evaluate whether agents can complete all in-scope work while avoiding out-of-scope actions, we build a challenging mobile benchmark. 
From a list of near identical entries, agents must act on every entry that satisfies the instruction and reject entries that violate its constraints.
We further introduce \textit{App-Level Progress} and \textit{Scope-Aware F1} to measure these two dimensions separately.
Experiments on AndroidWorld, MobileWorld, and our benchmark show that ATMem substantially improves long-horizon execution. 
With an 8B model, ATMem-UI achieves 76.6\% success rate on AndroidWorld, outperforming UI-TARS-2-230B by 3.3 percentage points and the same-sized MAI-UI by 5.9 points.
Our code and model will be publicly available.
\vspace{-1mm}
\end{abstract}
\abscontent

\section{Introduction}
\label{sec:intro}

Online mobile agents have made substantial progress in executing user instructions in real-world mobile environments, advancing from single-step grounding to multi-step navigation \citep{rawles2023androidinthewild, deng2024mobile, chen2024spa, yang2025probench, sumers2023cognitive}.
As tasks extend to long-horizon workflows \citep{kong2025mobileworld, rawles2024androidworld}, the agent must carry information across pages and applications while reading, verifying, modifying, and submitting it as the task unfolds.
In such workflows, the next action often depends on information observed, generated, or modified many steps earlier that is no longer visible on the current screen \citep{chai2025a3, chen2024spa, yang2025probench, yan2025stepguitechnicalreport}.

A natural response is to equip GUI agents with memory.
Existing approaches largely frame memory as a passive record of the past, whether through full interaction histories \citep{wang2024mobile, wang2024mobilev1}, trajectory summaries \citep{xiao2026ui, zhang2025appagent}, free-form notes \citep{wang2025mobile, ye2025mobile}, or retrieval over past traces \citep{shinn2023reflexion, park2023generative}.
Under this view, memory is framed as remembering more or retrieving better. 
We argue that this view is incomplete because making past information accessible does not necessarily make it useful for execution. 
As shown in Figure~\ref{fig:teaser}~(a), even when an agent retrieves the required value, it may still repeat a completed operation or omit a required one.
Record-centric memory preserves \textit{what was observed}, but does not bind each value to the subgoal it supports or indicate whether the value is pending, ready, or already used. 
The agent must therefore infer the current task state from accumulated records at every step, including which subgoals have been completed, which ones remain, and which action is currently valid. 
As workflows span more pages and applications, values become separated from the operations they are meant to guide, making this inference increasingly error-prone.

This raises a basic question that is rarely asked directly: \textcolor{tongyipurple}{\textit{\textbf{What memory do GUI agents actually need to drive long-horizon execution?}}}
Inspired by human cognition, where memory is not merely a passive record of past experience but an active mechanism for maintaining goals, tracking progress, and guiding future actions \citep{schacter2007cognitive, schacter2012constructive}, we argue that \textcolor{tongyipurple}{\textit{\textbf{GUI-agent memory should be an active task-driving state rather than an archive of past observations.}}}
Guided by this insight, we propose \textcolor{black}{\textbf{Active Task-Driving Memory (ATMem)}}, which is updated during interaction to track task progress and guide subsequent actions.
This view is particularly important in GUI workflows where task progress depends on the evolving status of task-relevant data rather than the mere availability of past observations.
As shown in Figure~\ref{fig:teaser}~(c), 83\% of AndroidWorld tasks involve data operations that advance the workflow by finding, verifying, modifying, or submitting task-relevant information.
In such workflows, each completed operation changes the execution state and determines what should be done next. 
For example, verifying that a contact satisfies a constraint can make it eligible for the next operation, while recording one recipe from a webpage can make the next recipe the current target for transfer to another app. 
Similarly, once a value has been submitted, the agent should mark it as used and avoid repeating the same operation.
These evolving data states determine which subgoals have been completed, which entries remain actionable, which constraints still apply, and which action is valid next. 
Memory for GUI agents should therefore maintain not only the values observed during execution, but also their provenance, current status, associated constraints, and role in task progress, so that memory can directly drive action selection throughout long-horizon execution.

ATMem operationalizes this idea by organizing task-relevant values, their structure, provenance, and current status into an active execution state that tracks completed subgoals, pending operations, and data available for the next action.
Unlike a scratchpad or a fixed-state tracker, ATMem is not a chronological note or a hand-engineered schema. 
It is induced from the task and updated by the agent during interaction.
ATMem maintains a hierarchical and extensible task state across app-level files and their underlying data units. 
\textbf{Workflow Progress} records the current phase and which app-level data files remain to be processed or have been completed. 
\textbf{Constraints} encode the instruction-defined logic and conditions independently of the schema. \textbf{Schema} defines the minimal task-relevant data units, such as fields, together with their properties. 
\textbf{ItemContent} stores the observed content and the execution status of individual data instances. 
Together, these sections identify remaining files, constraint-satisfying instances, and values available for the next operation. 
After each new observation or operation outcome, the agent updates ATMem and combines it with the current GUI observation and interaction history to select the next action.
This closed loop allows ATMem to externalize the agent's evolving execution state, helping it perceive workflow progress and maintain consistency across screens.

To equip the agent with ATMem, we train it in two stages. 
Supervised fine-tuning on verified trajectories teaches the agent to construct a valid ATMem structure, update it from new observations, and reference it when predicting actions. 
Given that this supervision specifies how ATMem should be formed but not whether relying on it improves task completion, we introduce \textbf{STR-GRPO}, a memory-aware online reinforcement learning method. 
During interaction, STR-GRPO learns an ATMem usage policy from task outcome rewards. 
It contrasts paired rollouts of the same task with ATMem enabled and disabled to estimate ATMem's contribution to task completion, while a memory cost-aware reward penalizes use that adds extra steps without improving the outcome. 
The agent therefore learns both how to maintain ATMem and when to rely on it for task completion.

To evaluate whether and how well agents can maintain progress and task scope over evolving task states, we build an online mobile benchmark for multi-entry tasks with scope constraints. 
Each task requires the agent to apply the instructed operation to every entry that satisfies the constraints while excluding entries with the same field structure that violate them. 
The benchmark holds the core operation schema fixed while varying the number of target entries and same-schema distractors. 
Difficulty therefore increases through coverage and scope control without introducing new operation types. 
It contains 32 manually designed templates. Using these templates, we construct three benchmark sets of increasing difficulty, each containing instances of all templates. 
Across these sets, the distractor-to-target ratio increases from 1.39$\times$ to 3.22$\times$.
Considering that the terminal success rate is binary and hides partial progress, we introduce two complementary metrics. 
\textit{App-Level Progress} measures how far the agent advances through the required operations in each application. 
\textit{Scope-Aware F1} measures precision and recall over task-relevant entries while penalizing operations on irrelevant ones.
Across standard benchmarks and our benchmark, ATMem and STR-GRPO yield the largest gains on workflows where progress is tightly coupled with evolving data states.

Our contributions are summarized as follows.

\begin{itemize}

    \item We propose \textbf{Active Task-Driving Memory (ATMem)}, which represents task-relevant data as a memory actively driving execution rather than an archive of past observations, enabling explicit tracking of data ownership, task role, task constraints, execution status, and workflow progress.
    \vspace{0.4em}
    \item We introduce \textbf{STR-GRPO}, a memory-aware online reinforcement learning method that turns ATMem usage into a learnable policy decision through memory-on and memory-off rollout interventions and memory-cost-aware optimization.
    \vspace{0.4em}
    \item We build a scalable data-centric online mobile-agent benchmark that evaluates cross-page, cross-application, and data-dependent workflows. It requires agents to complete all instruction-relevant operations while filtering same-schema distractors, and introduces \textit{App-Level Progress} to measure per-app operation completion and \textit{Scope-Aware F1} to evaluate target coverage under distractor filtering.
    \vspace{0.4em}
    \item We demonstrate state-of-the-art performance with our proposed agent \textbf{ATMem-UI}. It reaches 76.6\% success rate on AndroidWorld with an 8B model, outperforms UI-TARS-2-230B by 3.3 percentage points, and achieves a 36.2\% relative improvement over the strongest baseline on MobileWorld.

\end{itemize}

\section{Related Work}
\label{sec:related_work}

\noindent\textbf{Mobile GUI Agents.}
Mobile GUI agents have made rapid progress in recent years \citep{zhou2025mai, xu2026mobile, qin2025ui, wang2025ui, gu2025ui, gelab_zero_2025}. 
AutoDroid \citep{wen2024autodroid} formulates Android task automation as a multimodal interaction problem, and subsequent systems such as AppAgent \citep{zhang2025appagent}, AppAgent-v2 \citep{li2024appagent}, MAI-UI \citep{zhou2025mai}, and Mobile-Agent-v3/v3.5 \citep{xu2026mobile, ye2025mobile} extend agent capabilities to more complex smartphone and GUI tasks. 
In parallel, AndroidWorld \citep{rawles2024androidworld}, AndroidLab \citep{xu2025androidlab}, and MobileWorld \citep{kong2025mobileworld} provide realistic online environments with programmatic evaluation. 
More recent works, such as MobileGUI-RL \citep{shi2025mobilegui} and MobileRL \citep{xu2025mobilerl}, explore online reinforcement learning to improve long-horizon execution, while AgentProg \citep{tian2025agentprog} reduces reliance on raw history replay through programmatic context management and belief-state tracking. 
Together, these efforts have substantially advanced GUI interaction capabilities, online benchmarks, and long-horizon mobile-agent research.

\noindent\textbf{Memory for GUI and Mobile Agents.}
Memory in mobile and GUI agents has developed along both experience-oriented and execution-oriented directions \citep{li2024appagent, wen2024autodroid, lee2024mobilegpt, zhu2025moba, ye2025mobile, wang2025mobile, sun2025fairy, sun2026magnet, xiao2026ui, liu2026memgui, zhu2026hybrid, yu2026graphpilot,cheng2025mga, shi2026androtmem, li2025echotrail, chen2026skilldroid}. 
Experience-oriented methods store app knowledge \citep{li2024appagent}, human-like app memory \citep{wen2024autodroid, lee2024mobilegpt}, past trajectories \citep{zhu2025moba}, skills \citep{wang2025mobile, sun2025fairy}, or interaction patterns \citep{yu2026graphpilot} to support knowledge reuse, generalization, and interface adaptation. 
Execution-oriented methods, such as note-taking \citep{wang2025mobile}, anchored intermediate states \citep{zhu2026hybrid}, and program-based context management \citep{xiao2026ui}, preserve key information or compact task context \citep{sun2026magnet, shi2026androtmem} from ongoing trajectories to support subsequent actions.
While these works highlight the importance of memory for long-horizon GUI execution, they mainly represent memory as reusable experience, recorded notes, retrieved context, or program/interface state. 
In contrast, our work explicitly models task-relevant data objects, their structures, values, and execution states, and uses them as actionable signals to guide execution.

\begin{figure*}[t]
\centering
\includegraphics[width=1.0\textwidth]{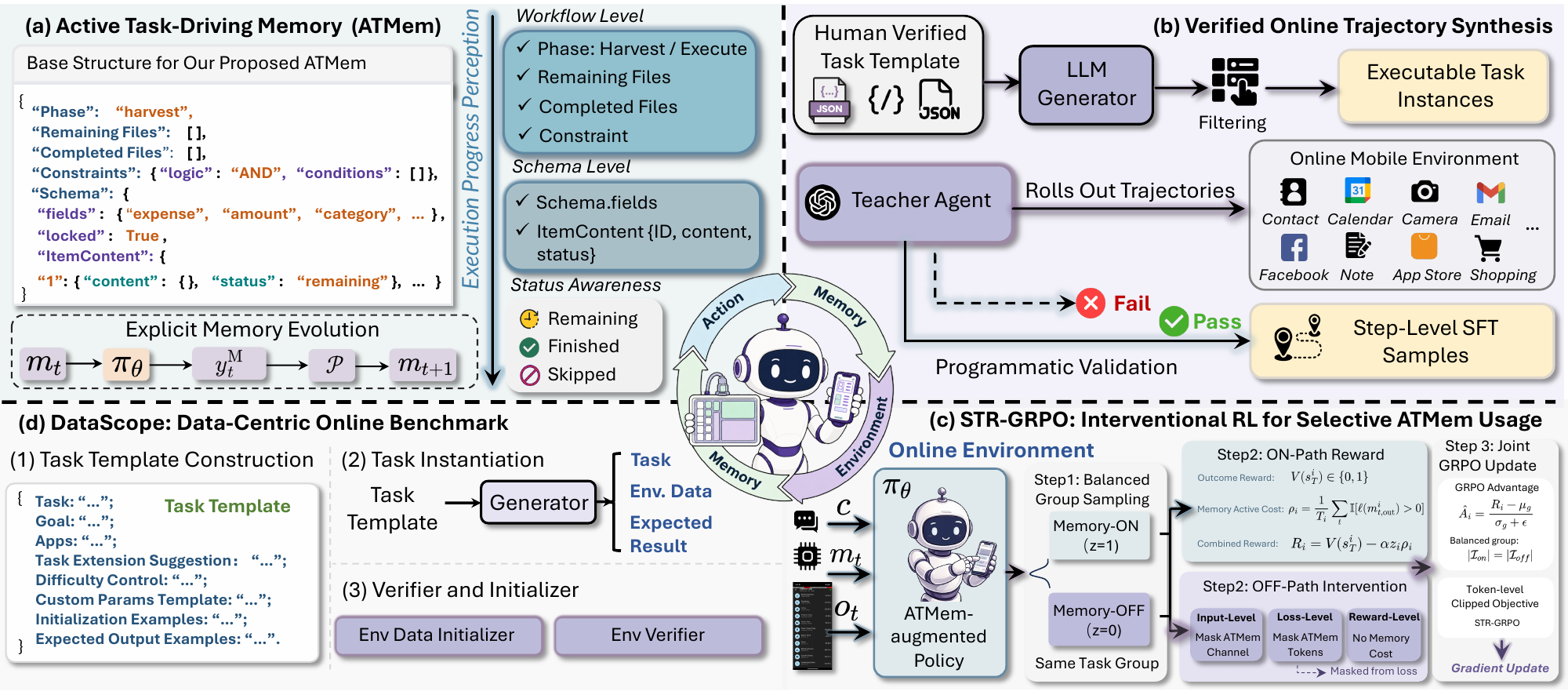}
\vspace{-1.5em}
\caption{
\textbf{Overview of our methodology.}
(a) ATMem organizes task-relevant data into a structured execution state with constraints, schema fields, item content, and item-level status.
(b) Verified SFT data are synthesized through task-template instantiation, teacher-agent rollouts, and environment validation.
(c) STR-GRPO uses balanced memory-ON/OFF interventions to estimate ATMem utility and learn selective memory invocation.
(d) DataScope evaluates data-centric workflows built from controlled target entries, same-schema distractors, and calibrated task verifiers.
}
\label{fig:overview}
\end{figure*}

\section{Methods}
\label{sec:method}

As shown in Figure~\ref{fig:overview}, our framework consists of four components. 
We first introduce ATMem, which turns task-relevant data from passive interaction records into structured execution states that agents actively maintain and use to guide actions across interaction steps (Section~\ref{sec:SDTM_Structure}).
We then describe verified online trajectory synthesis to build high-quality SFT data (Section~\ref{sec:traj_syn}). 
Next, we present STR-GRPO for improving memory-conditioned execution (Section~\ref{sec:RLLearning}). 
Finally, we detail the construction of our DataScope benchmark for evaluating long-horizon tasks (Section~\ref{sec:benchmark}).

\subsection{ATMem: Active Task-Driving Memory}
\label{sec:SDTM_Structure}

\noindent\textbf{From Passive History to Active Task-Driving Memory.}
A standard mobile GUI agent can be formulated as a history-conditioned action policy.
At step $t$, given the task instruction $c$, the current GUI observation $o_t$, and the interaction history $h_{<t}$, the agent generates a textual response $y_t$, from which an output parser $\mathcal{P}$ extracts an executable action $a_t$ and an intermediate reasoning trace $\rho_t$:
\begin{equation}
y_t \sim \pi_\theta(\cdot \mid c, o_t, h_{<t}), \qquad
(\rho_t, a_t) = \mathcal{P}(y_t).
\end{equation}
Although $h_{<t}$ preserves values observed or produced in previous steps, it stores them as chronological records rather than as a task state organized around the data being processed. 
The agent must therefore repeatedly infer which item a value refers to, whether it satisfies the task constraints, and whether the associated work has already been completed. 
As trajectories grow longer and contain similar fields, repeated values, distractors, or outdated states, inferring task progress from accumulated records becomes increasingly unreliable, often leading to repeated or missed operations. This motivates a shift from passive interaction history to memory actively driving execution, where task progress, data eligibility, and remaining operations are explicitly maintained.

Our key insight is that many long-horizon GUI workflows require explicit tracking of the execution status of task-relevant data. 
We therefore introduce ATMem, an active task-driving memory that organizes task-relevant information as a hierarchical execution state.
ATMem operationalizes this shift through a two-level structure that turns memory from a chronological record into an agent-maintained execution state.
Formally, ATMem maintains a memory state $m_t = (\phi_t, E_t)$, where $\phi_t$ summarizes workflow-level progress and task constraints, and $E_t = \{e_i\}$ is a dynamic set of task-relevant item entries.
Each entry $e_i$ stores a stable item identifier, structured task-critical content, and an execution status $s_i \in \{\texttt{remaining}, \texttt{finished}, \texttt{skipped}\}$, indicating whether the item still needs to be processed, has been completed, or should be skipped.
ATMem therefore represents not only the available data, but also the item-level state needed to determine remaining work.

Before each model call, the current ATMem state $m_t$ is inserted into the system prompt as the current task-driving memory.
Conditioned on this injected memory, the agent generates a memory-augmented response $y_t^{\mathrm{M}}$.
The parser $\mathcal{P}$ then extracts the updated task-driving memory $m_{t+1}$, the intermediate reasoning trace $\rho_t$, and the executable GUI action $a_t$ from the model response:
\begin{equation}
y_t^{\mathrm{M}} \sim \pi_\theta(\cdot \mid c, o_t, h_{<t}, m_t), 
\qquad
(m_{t+1}, \rho_t, a_t) = \mathcal{P}(y_t^{\mathrm{M}}).
\end{equation}
The extracted $m_{t+1}$ is then inserted into the next system prompt as the current memory, enabling the agent to carry forward an up-to-date execution record without relying on history re-examination.

\noindent\textbf{Basic Structure Design.}
As illustrated in Figure~\ref{fig:overview} (a), ATMem organizes memory into two complementary levels that implement the shift from passive history to active task-driving state.
Workflow-level fields encode where the agent is in the task and what global constraints should guide execution, while schema-level fields define and track the minimal actionable data units on which GUI operations are performed.
A minimal actionable data unit is the smallest semantically complete data object that can be independently operated on during task execution.
For example, an expense record comprising \texttt{expense}, \texttt{amount}, and \texttt{category} forms one such unit, whereas any individual attribute alone is insufficient to constitute an operable target.

\noindent\textbf{\ding{172} Workflow-level fields.}
Workflow-level fields capture task-level execution progress and global task conditions with a fixed structure shared across tasks.
The \texttt{Phase} field partitions execution into two macro-stages, \texttt{harvest} and \texttt{execute}.
During \texttt{harvest}, the agent inspects available sources and populates task-relevant item entries.
During \texttt{execute}, the agent performs data-dependent operations over the populated entries.
The phase transitions from \texttt{harvest} to \texttt{execute} once sufficient items have been collected and no sources remain to be inspected.
For tasks that require source-level tracking, \texttt{RemainingFiles} and \texttt{CompletedFiles} maintain a checklist of sources pending inspection and sources already processed, providing the concrete signal for phase transition.
When source-level tracking is not required, both fields remain empty.
The \texttt{Constraints} field stores task-derived filtering conditions as global execution rules, determining which collected items are eligible for operation, such as selecting only records that satisfy a specified date range, category, or recipient.

\noindent\textbf{\ding{173} Schema-level fields.}
Schema-level fields define the minimal actionable data units for a given task and maintain their execution states.
\texttt{Schema.fields} specifies the attribute set that constitutes one semantically complete data unit, such as \texttt{expense}, \texttt{amount}, \texttt{category} in an expense-management task or \texttt{recipient}, \texttt{subject} in an email task.
During schema construction, explicit task-specific attributes can be added when required.
Once the relevant attributes are identified, \texttt{Schema.locked} freezes the field set to prevent structural drift across steps.
\texttt{Schema.ItemContent} enumerates instantiated data units, each indexed by a stable identifier and associated with structured field values and an execution status from $\{\texttt{remaining}, \texttt{finished}, \texttt{skipped}\}$.
Here, \texttt{remaining} and \texttt{finished} indicate whether a unit still awaits processing or has been completed.
\texttt{skipped} provides an item-level refinement complementary to \texttt{Constraints}.
While \texttt{Constraints} captures global eligibility rules derived from the task instruction, \texttt{skipped} allows the agent to mark a specific item as inapplicable when richer item-level details become available during execution.

Together, workflow-level fields control task flow and global filtering conditions, while schema-level fields define and track the concrete data units to be operated on.
The base ATMem structure provides a shared template across tasks, while task-specific fields can be added to accommodate different workflows, such as date attributes for calendar tasks.
We provide the execution prompt in Appendix~\ref{app:agent_execution_prompt}.

\subsection{Verified Online Rollout Synthesis.}
\label{sec:traj_syn}
As shown in Figure~\ref{fig:overview} (b), we synthesize supervised data through verifier-gated online rollouts in a virtual mobile environment with 20 applications.
We construct 120 task templates, each human-designed and verified, to define a controllable and verifiable task space rather than treating tasks as isolated natural-language instructions.
Each template consists of two executable components and a set of instantiation guidelines.
The initializer constructs the required environment state by injecting task-relevant data and distractors, and the task-specific verifier determines whether the terminal environment state satisfies the intended goal.
The guidelines include a goal pattern, descriptions of required data objects and fields, task-instance examples, and suggested expansion directions to support downstream instantiation.

Based on these templates, an LLM-assisted generator instantiates about 1.2K candidate task instances with user-facing instructions and concrete initialization data.
The generator leverages template-provided examples and expansion directions to produce valid task variations, while executability and correctness are determined by the initializer and verifier.
After removing unreachable, malformed, or invalid instances, we obtain 1.1K executable task instances for rollout collection.

For each executable task instance, we run an online rollout framework that combines an LLM planner \citep{openai2025introducing} with UIIns \citep{chen2025ui}, a UI grounding model that maps predicted operations to executable GUI actions.
The planner is instructed to maintain ATMem only when structured memory is needed, such as in workflows involving multiple data items, cross-file information collection, or repeated data-dependent operations.
When ATMem is not needed, the memory state is kept empty.
At each step, the rollout framework produces a response containing the updated memory state, an intermediate reasoning trace, and an executable GUI action.
A trajectory is retained only if its terminal environment state passes the corresponding task verifier.
This verifier-gated process yields 21,713 step-level SFT samples.
Each sample maps the current execution context $(c,o_t,h_{<t},m_t)$ to the target response $y_t=(m_{t+1},\rho_t,a_t)$, where $m_t$ and $m_{t+1}$ are empty when ATMem is not activated.

\subsection{Interventional Online RL for Effective ATMem Usage}
\label{sec:RLLearning}

\noindent\textbf{Online RL Setup.}
After SFT on verifier-accepted trajectories, the agent can construct and update ATMem during GUI execution.
However, SFT only imitates memory patterns observed in successful rollouts and does not estimate whether ATMem contributes to the current policy's execution.
We therefore introduce Structured Task-driven Reward GRPO (STR-GRPO), an interventional online RL method.
During paired rollouts, we intervene on the explicit ATMem channel by enabling or masking memory access while keeping the task initialization unchanged.
The resulting rewards provide a relative estimate of ATMem's execution utility.

\noindent\textbf{Memory-Conditional Intervention.}
Given a task $q$, STR-GRPO samples a group of $N_g$ rollouts from the current policy and assigns each rollout to one of two memory conditions.
Let $z_i \in \{0,1\}$ denote the intervention for rollout $i$, where $z_i=1$ enables the explicit ATMem channel and $z_i=0$ masks it.
At step $t$, the policy input can be formulated as:
\begin{equation}
x_t^{(z_i)} = (c, o_t, h_{<t}, \tilde{m}_t^{(z_i)}),
\qquad
\tilde{m}_t^{(z_i)} =
\begin{cases}
m_t, & z_i=1,\\
\varnothing, & z_i=0.
\end{cases}
\end{equation}

Here, $c$ is the task instruction, $o_t$ is the current observation, and $h_{<t}$ denotes the standard interaction history.
The intervention changes only the explicit memory channel, while previous observations, executed actions, and reasoning traces remain available in both conditions.
To reduce the effect of task difficulty, we apply a balanced assignment within each group, with $N_g/2$ rollouts assigned to each condition.

\noindent\textbf{Reward Design and STR-GRPO Objective.}
Given the balanced memory intervention above, STR-GRPO assigns each rollout a trajectory-level reward that combines terminal task validation and a memory-active cost:
\begin{equation}
R_i = V(s^i_{T_i}) - \alpha z_i \rho_i, \qquad
\rho_i = \frac{1}{T_i} \sum_{t=1}^{T_i}
\mathbb{I}\left[\ell({m}^i_{t,\mathrm{out}})>0\right].
\end{equation}
Here $V(s^i_{T_i})\in\{0,1\}$ is the verifier reward for the terminal environment state, ${m}^i_{t,\mathrm{out}}$ denotes the ATMem block generated at step $t$, and $\ell(m)=0$ if and only if all fields of the ATMem block are empty.
Thus, $\rho_i$ measures the fraction of memory-active steps, and the gate $z_i$ restricts the memory cost to memory-ON rollouts, where ATMem is propagated across turns.

For each rollout group $g$, we compute the group-normalized advantage over the balanced mixture of memory-ON and memory-OFF trajectories $\hat{A}_i = (R_i - \mu_g) / (\sigma_g + \epsilon_A)$, where $\mu_g$ and $\sigma_g$ are the mean and standard deviation of rewards within group $g$, and $\epsilon_A$ is a small constant for numerical stability.
This provides a shared baseline for both memory conditions within the same task group, so that advantages directly reflect the marginal utility of the ATMem channel.

Overall, the STR-GRPO objective is defined as:
\begin{equation}
\mathcal{L}(\theta) = -\mathbb{E}\!\left[\sum_{i,t,k}
\min\!\left(r_{t,k}\,\hat{A}_i,\;
\mathrm{clip}(r_{t,k},1\pm\varepsilon)\,\hat{A}_i\right)\right],
\quad
r_{t,k} = \frac{\pi_{\theta}(y^i_{t,k}\mid x^i_t,y^i_{t,<k})}
{\pi_{\theta_{\mathrm{old}}}(y^i_{t,k}\mid x^i_t,y^i_{t,<k})}.
\end{equation}
For memory-ON rollouts, the loss covers the ATMem block, reasoning trace, and executable action.
For memory-OFF rollouts, ATMem block tokens are masked from the loss, and the loss covers only the reasoning trace and executable action under an empty memory channel.
Memory-ON rollouts receive higher relative advantages only when ATMem improves verifier reward enough to offset the memory-active cost, discouraging redundant memory dependence and promoting selective structured memory usage.

\begin{figure}
\vspace{-0.8em}
\centering
\includegraphics[width=0.75\linewidth]{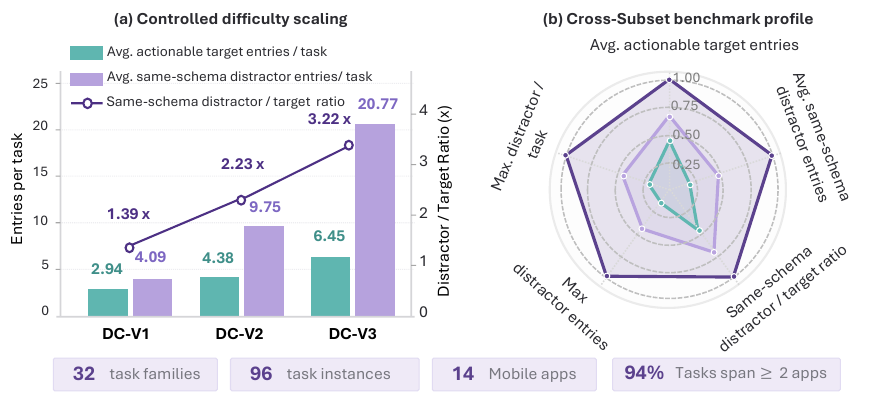}
\caption{\textbf{DataScope statistics and controlled difficulty scaling.}
\textbf{(a)} Bars show the average numbers of actionable target entries and same-schema distractors per task from DC-V1 to DC-V3, while the line shows the distractor-to-target ratio.
The controlled increase in both quantities raises the ratio from $1.39\times$ to $3.22\times$, testing whether agents can maintain target coverage while filtering increasingly confusable data within the same task families.
\textbf{(b)} Radar profiles summarize average target and distractor counts, their ratio, and the maximum distractor load and ratio across difficulty levels.
The outward shift from DC-V1 to DC-V3 indicates progressively greater data-coverage and distractor-filtering demands.
DataScope contains 32 task families, 96 task instances, and 14 mobile apps, with 94\% of tasks spanning at least two apps.
}
\label{fig:benchmark_stats}
\vspace{-0.4em}
\end{figure}

\subsection{DataScope Online Mobile Benchmark and Metrics}
\label{sec:benchmark}
\noindent\textbf{Benchmark Overview.}
We introduce DataScope, an online mobile benchmark for multi-entry tasks with controlled scope difficulty. 
Each task requires the agent to apply the instructed operations to every entry that satisfies the instruction constraints while excluding structurally matched distractors that share the same field schema or data type. 
Unlike prior benchmarks that evaluate end-to-end GUI task completion, DataScope controls both target coverage and distractor filtering while holding the underlying operation schema fixed.

As shown in Figure~\ref{fig:benchmark_stats}, DataScope contains 32 manually designed task templates, each instantiated at three difficulty levels from DC-V1 to DC-V3. 
The difficulty levels increase the number of target entries and structurally matched distractors without introducing new operation types. 
The distractor-to-target ratio increases from $1.39\times$ in DC-V1 to $3.22\times$ in DC-V3, requiring broader target coverage and more precise distractor filtering. 
In total, DataScope includes 96 evaluation instances across 14 apps, with 94\% of tasks spanning at least two apps.

\noindent\textbf{Task Construction.}
Each task template is defined by a human-verified workflow specification:
\begin{equation}
\Omega =
\left(
\mathcal{A},
g_0,
\mathcal{X},
\mathcal{C}_{\ell},
I_{\Omega},
\mathcal{S}_{\mathrm{exp}},
\mathcal{B}_{\mathrm{val}}
\right).
\end{equation}
Here $\mathcal{A}$ specifies the apps involved in the workflow, and $g_0$ defines the task objective to be completed by the agent. 
$\mathcal{X}$ describes the structure and seed examples of task data, including target entries that must be operated on and confusable distractors that must remain unchanged. 
$\mathcal{C}_{\ell}$ defines the data and operation constraints for difficulty level $\ell$. The initializer $I_{\Omega}$ writes an instantiated data sample into the corresponding app storage. $\mathcal{S}_{\mathrm{exp}}$ specifies the terminal-state requirements for successful execution, while $\mathcal{B}_{\mathrm{val}}$ contains controlled environment states and their expected evaluation scores for verifier calibration.

Our task construction proceeds in three stages.
\textbf{First}, the workflow specification $\Omega$ is manually constructed and verified.
\textbf{Second}, an LLM-assisted generator instantiates executable task instances from $\Omega$.
Each instance consists of a task instruction $c$, task-specific initialization data $x$, and a terminal-state specification $y^\star$.
The initializer $I_{\Omega}$ writes $x$, including target entries and confusable distractors, into the corresponding app storage to construct the initial environment state.
The agent must therefore access and manipulate all task-relevant data through real GUI interaction.
\textbf{Third}, we generate a functional verifier for each task template. 
The LLM framework synthesizes the verifier from the initialization function, terminal-state requirements, and validation basis. 
For calibration, $\mathcal{B}_{\mathrm{val}}$ provides controlled environment states covering successful completion, missing required operations, over-operation on distractors, missing app-level operations, and incorrect field values. 
Each state is paired with its expected terminal success, App-Prog., and Scope-F1 scores. 
A generated verifier is validated against all controlled states. 
If its outputs do not match the expected scores, the verifier is manually corrected and revalidated until it passes all calibration cases.
Further construction details and template examples are provided in Appendix~\ref{app:benchmark_details}, ~\ref{app:env_init_verifier}, and ~\ref{app:workflow_template_example}.

\noindent\textbf{App- and Data-Level Metrics.}
Terminal success equals one only when all task-specific terminal-state requirements are satisfied.
Although appropriate for measuring complete execution, it assigns the same zero score to unsuccessful trajectories that terminate at different stages or exhibit different data-operation errors. 
We therefore introduce two complementary metrics at finer levels of granularity. 
App-Prog measures the completion of app-level subgoals, while Scope-F1 evaluates the completeness of individual data operations.

For task $q$, let $\mathcal{A}_q$ be the set of apps involved, and let $c_{q,a} \in \{0,1\}$ indicate whether all required operations in app $a$ are completed.
\textit{App-Prog.} measures the fraction of apps with all required operations satisfied:
\begin{equation}
S_{\mathrm{prog}}(q)
= \frac{1}{|\mathcal{A}_q|}
\sum_{a \in \mathcal{A}_q} c_{q,a}.
\end{equation}
An app contributes only when its complete required subgoal is achieved.
App-Prog therefore provides coarse-grained progress awareness across a multi-app workflow and distinguishes trajectories that complete different numbers of required applications.
It focuses only on required subgoal completion, while unintended operations are evaluated separately by Scope-F1.

\textit{Scope-F1} evaluates execution at the atomic data-unit level.
For each app $a$, let $\mathcal{D}^{\star}_{q,a}$ denote the set of required atomic data units and $\hat{\mathcal{D}}_{q,a}$ the set of data units correctly handled by the agent.
Each unit is the smallest independently verifiable element of task execution, jointly identified by its data entry, field, and expected value where applicable.
A unit is counted as correctly handled only when the final environment state reflects the intended change on the intended entry.
We define
$\mathrm{TP}_{q,a} = \left|\hat{\mathcal{D}}_{q,a} \cap 
\mathcal{D}^{\star}_{q,a}\right|,
\quad
\mathrm{FP}_{q,a} = \left|\hat{\mathcal{D}}_{q,a} \setminus 
\mathcal{D}^{\star}_{q,a}\right|,
\quad
\mathrm{FN}_{q,a} = \left|\mathcal{D}^{\star}_{q,a} \setminus 
\hat{\mathcal{D}}_{q,a}\right|,
$
and compute the per-app F1 as
$F_{q,a} = 2\,\mathrm{TP}_{q,a} \;/\; 
(2\,\mathrm{TP}_{q,a} + \mathrm{FP}_{q,a} + \mathrm{FN}_{q,a} + 
\epsilon)$,
where $\epsilon$ is a small constant for numerical stability.
Let $\mathcal{A}^{\mathrm{act}}_q$ denote the set of apps containing at least one data unit modified by the agent.
Scope-F1 is macro-averaged over both required and additionally modified apps:
\begin{equation}
S_{\mathrm{scope}}(q)
= \frac{1}{\left|\mathcal{A}_q \cup 
  \mathcal{A}^{\mathrm{act}}_q\right|}
  \sum_{a \in \mathcal{A}_q \cup \mathcal{A}^{\mathrm{act}}_q} 
  F_{q,a}.
\end{equation}

Missing required units are counted as false negatives. 
Distractor entries, incorrect fields, incorrect values, and changes in unrelated apps are counted as false positives. 
Including additionally modified apps ensures that data changes outside the required workflow are also penalized.

Terminal success, App-Prog, and Scope-F1 evaluate execution at the task, application, and atomic data-unit levels, respectively. 
App-Prog indicates how far the agent progresses across the required applications, while Scope-F1 measures whether the required data units are correctly completed without modifying data outside the intended scope.

\section{Experiments}
\label{sec:exp}

\subsection{Experimental Setup}
\noindent\textbf{Training Environment and Infrastructure.}
We use a virtual Android platform containing 20 applications for both SFT trajectory collection and online RL training. 
SFT trajectories are retained only after successful task-specific verification. 
For online RL, we build an asynchronous on-policy training arena with 128 active containerized Android virtual devices distributed across multiple bare-metal servers. 
Multi-turn rollouts are generated by the latest policy and asynchronously collected to reduce idle time.

Policy optimization is performed on 128 NVIDIA H20 GPUs using hybrid parallelism to support long multimodal trajectories containing GUI observations, reasoning traces, actions, and ATMem states. 
This infrastructure enables scalable rollout collection and end-to-end optimization for long-horizon mobile interaction. 
Further implementation details and hyperparameters are provided in Appendix~\ref{app:online_rl_infra} and Appendix~\ref{app:training_infra}.

\noindent\textbf{Evaluation Benchmarks.}
We evaluate our models on AndroidWorld and MobileWorld, two widely used online mobile-agent benchmarks, together with our proposed DataScope benchmark. 
AndroidWorld and MobileWorld evaluate general online mobile-agent performance, while DataScope focuses on cross-app workflows that require target data identification, distractor filtering, and stateful data manipulation.

DataScope contains three difficulty levels, instantiated from the same 32 task families by progressively increasing target entries, confusable distractors, and required operations. 
Each level contains 32 task instances, yielding 96 instances across 14 mobile apps. Overall, 94\% of DataScope tasks involve at least two apps.

\noindent\textbf{Baselines.}
Our 4B and 8B models are initialized from Qwen3-VL-4B and Qwen3-VL-8B, respectively~\citep{Qwen3-VL}. We compare them with the corresponding Qwen3-VL backbones and representative mobile agents, including GUI-Owl~\citep{xu2026mobile,ye2025mobile}, UI-Venus~\citep{gu2025ui}, Doubao-1.5-UI-TARS~\citep{qin2025ui}, and GELab-Zero~\citep{gelab_zero_2025}.

We additionally report a strong teacher-agent reference that uses GPT-5~\citep{openai2025introducing} for high-level planning and UIIns~\citep{chen2025ui} for visual grounding. 
Because this system relies on a substantially stronger proprietary planner, we treat it as a teacher reference rather than a size-matched baseline.

\begin{table*}[!t]
\centering
\scriptsize
\setlength{\tabcolsep}{3pt}

\begin{minipage}[t]{0.49\linewidth}
\centering
\renewcommand{\arraystretch}{1.0}
\caption{
Performance on \textbf{AndroidWorld}.
SR denotes success rate.
\besttext{Green}/\secondtext{purple} indicates the best/second-best results within the same scale group.
}
\vspace{-0.8em}
\label{tab:androidworld}
\begin{tabularx}{\linewidth}{@{}>{\raggedright\arraybackslash}Xcc@{}}
\toprule
\textbf{Method} & \textbf{Params} & \textbf{SR} \\
\midrule
\rowcolor{gray!12}
\multicolumn{3}{@{}l@{}}{\textbf{\textit{Larger and proprietary models for reference}}} \\
\refcell{UI-TARS-SFT} \citep{qin2025ui} & \refcell{72B} & \refcell{65.9} \\
\refcell{UI-Venus} \citep{gu2025ui} & \refcell{72B} & \refcell{65.9} \\
\refcell{Qwen3-VL-235B-A22B} \citep{Qwen3-VL} & \refcell{235B} & \refcell{63.7} \\
\refcell{UI-TARS-1.5} \citep{ui-tars-15-seed} & \refcell{--} & \refcell{64.2} \\
\refcell{Gemini-2.5-Pro} \citep{geminicom} & \refcell{--} & \refcell{69.7} \\
\refcell{Seed1.8} \citep{seed18} & \refcell{--} & \refcell{70.7} \\
\refcell{UI-TARS-2} \citep{wang2025ui} & \refcell{230B} & \refcell{73.3} \\
\midrule
\rowcolor{gray!12}
\multicolumn{3}{@{}l@{}}{\textbf{\textit{Recent 4B/8B models}}} \\
Step-GUI \citep{yan2025stepguitechnicalreport} & 8B & 67.7 \\
MAI-UI \citep{zhou2025mai} & 8B & 70.7 \\
GUI-Owl-1.5-Thinking \citep{xu2026mobile} & 8B & \second{71.6} \\
\rowcolor{gray!6}
\textbf{ATMem-UI} (\textcolor{tongyipurple}{Ours})  & 4B & 65.5 \\
\rowcolor{gray!6}
\textbf{ATMem-UI} (\textcolor{tongyipurple}{Ours})  & 8B & \best{76.6} \\
\bottomrule
\end{tabularx}
\end{minipage}
\vspace{-1.0em}
\hfill
\begin{minipage}[t]{0.49\linewidth}
\centering
\renewcommand{\arraystretch}{1.37}
\caption{
Generalization on \textbf{MobileWorld} in unseen environments.
\besttext{Green}/\secondtext{purple} indicates the best/second-best results within the same scale group.
}
\vspace{-0.8em}
\label{tab:mobileworld}
\begin{tabularx}{\linewidth}{@{}>{\raggedright\arraybackslash}Xcc@{}}
\toprule
\textbf{Method} & \textbf{Params} & \textbf{SR} \\
\midrule
\rowcolor{gray!12}
\multicolumn{3}{@{}l@{}}{\textbf{\textit{Larger models for reference}}} \\
\refcell{UI-Venus-1.5} \citep{team2026ui} & \refcell{30B} & \refcell{17.1} \\
\refcell{UI-Venus} \citep{gu2025ui} & \refcell{72B} & \refcell{16.4} \\
\refcell{GUI-Owl} \citep{ye2025mobile} & \refcell{72B} & \refcell{8.5} \\
\midrule
\rowcolor{gray!12}
\multicolumn{3}{@{}l@{}}{\textbf{\textit{Recent 4B/7B/8B models}}} \\
GUI-Owl \citep{ye2025mobile} & 7B & 7.7 \\
UI-Venus \citep{gu2025ui} & 7B & 8.5 \\
GELab-Zero \citep{gelab_zero_2025} & 4B & 16.1 \\
\rowcolor{gray!6}
\textbf{ATMem-UI} (\textcolor{tongyipurple}{Ours})  & 4B & \second{20.5} \\
\rowcolor{gray!6}
\textbf{ATMem-UI} (\textcolor{tongyipurple}{Ours}) & 8B & \best{23.3} \\
\bottomrule
\end{tabularx}
\end{minipage}\
\vspace{-1.5em}
\end{table*}

\subsection{Main Results}
\noindent\textbf{Performance on General Online Benchmarks.}
As shown in Table~\ref{tab:androidworld}, ATMem-UI-8B achieves 76.6\% SR on AndroidWorld, outperforming all recent 4B/8B agents by a clear margin, with gains of $+$5.0 points over GUI-Owl-1.5-Thinking and $+$5.9 points over MAI-UI-8B.
More tellingly, it surpasses UI-TARS-2-230B by 3.3 points while using roughly $1/29$ of its parameters.
This disproportionate gain suggests that the bottleneck in long-horizon mobile execution is not only model capacity.

The 4B results further illustrate the scale efficiency of structured memory.
ATMem-UI-4B achieves 65.5\% SR, within 0.4 points of the 72B-scale UI-TARS-SFT and UI-Venus, and exceeds Qwen3-VL-235B-A22B.
This comparison suggests that explicit state tracking can compensate for part of the capability gap usually addressed by parameter scaling, especially in long-horizon tasks where failures often arise from 
losing intermediate task state rather than from single-step perception or action selection alone.

Table~\ref{tab:mobileworld} evaluates cross-environment generalization on MobileWorld, where all environments are unseen during training.
ATMem-UI-8B achieves 23.3\% SR, outperforming the strongest non-ours baseline, GELab-Zero-4B, by 7.2 points and exceeding the 30B reference, UI-Venus-1.5, by 6.2 points.
The 4B variant already surpasses all listed non-ours baselines, including 72B references.
This strong transfer suggests that ATMem does not merely memorize training-app interaction patterns; instead, it provides a more portable representation of task progress and task-relevant data state, enabling better generalization to new mobile applications and workflows.

\noindent\textbf{Performance on our benchmark.}
DataScope directly probes the capabilities ATMem is designed to support, including tracking target data items across sources and maintaining execution scope over long cross-app workflows.
As shown in Table~\ref{tab:data_centric_results}, existing end-to-end agents struggle substantially on all difficulty levels.

On DC-V1, the strongest end-to-end baseline MAI-UI-8B obtains only 3.1\% SR, 8.8\% $S_{\mathrm{prog}}$, and 10.7\% $S_{\mathrm{scope}}$.
ATMem-UI-8B improves these to 6.2\%, 11.7\%, and 15.7\%, yielding gains of $+$3.1, $+$2.9, and $+$5.0 points, respectively.
The larger gain on $S_{\mathrm{scope}}$ is especially informative: it indicates that ATMem improves not only whether the agent makes progress, but also whether it operates on the correct set of required 
data items.
This is precisely the failure mode targeted by ATMem's explicit \texttt{Constraints} field and item-level status tracking, which provide a persistent record of which data objects remain eligible, completed, or skipped.

\begin{table}[t]
\centering
\caption{
Performance on our benchmarks.
We additionally provide a strong agentic framework as a competitive reference baseline.
SR denotes success rate, $S_{\mathrm{prog}}$ denotes progress score, and $S_{\mathrm{scope}}$ denotes scope score.
\besttext{Green}/\secondtext{purple} indicates the best/second-best results among end-to-end mobile agents.
}
\vspace{-0.8em}
\label{tab:data_centric_results}
\scriptsize
\setlength{\tabcolsep}{4.5 pt}
\renewcommand{\arraystretch}{1.1}

\resizebox{1\linewidth}{!}{
\begin{tabular}{@{}lcccccccccc@{}}
\toprule
\textbf{Method} & \textbf{Params}
& \multicolumn{3}{c}{\textbf{Data-Scope-V1}}
& \multicolumn{3}{c}{\textbf{Data-Scope-V2}}
& \multicolumn{3}{c}{\textbf{Data-Scope-V3}} \\
\cmidrule(lr){3-5}
\cmidrule(lr){6-8}
\cmidrule(l){9-11}
&
& \textbf{SR} & $\boldsymbol{S_{\mathrm{prog}}}$ & $\boldsymbol{S_{\mathrm{scope}}}$
& \textbf{SR} & $\boldsymbol{S_{\mathrm{prog}}}$ & $\boldsymbol{S_{\mathrm{scope}}}$
& \textbf{SR} & $\boldsymbol{S_{\mathrm{prog}}}$ & $\boldsymbol{S_{\mathrm{scope}}}$ \\
\midrule

\rowcolor{gray!12}
\multicolumn{11}{@{}l@{}}{\textbf{\textit{Agentic framework for reference}}} \\

\refcell{GPT-5.2+ UIIns \citep{openai2025introducing, chen2025ui}}
& \refcell{--}
& \refcell{18.8} & \refcell{35.4} & \refcell{38.4}
& \refcell{6.2} & \refcell{16.1} & \refcell{21.3}
& \refcell{3.1} & \refcell{9.3} & \refcell{13.2} \\

\midrule

\rowcolor{gray!12}
\multicolumn{11}{@{}l@{}}{\textbf{\textit{End-to-end mobile agents}}} \\

GELab-Zero \citep{gelab_zero_2025}
& 4B
& 0.0 & 4.2 & 3.9
& 0.0 & 3.6 & 6.4
& 0.0 & 2.6 & 4.4 \\

UI-TARS-1.5 \citep{ui-tars-15-seed}
& 7B
& 0.0 & 1.6 & 1.7
& 0.0 & 1.6 & 1.8
& 0.0 & 1.6 & 1.5 \\

GUI-Owl \citep{ye2025mobile}
& 7B
& 0.0 & 3.6 & 5.7
& 0.0 & 1.6 & 1.1
& 0.0 & 1.6 & 1.5 \\

UI-Venus \citep{gu2025ui}
& 7B
& 0.0 & 1.6 & 3.8
& 0.0 & 1.6 & 0.6
& 0.0 & 0.0 & 0.7 \\

GUI-Owl-1.5 \citep{xu2026mobile}
& 8B
& \second{3.1} & 4.7 & 7.3
& 0.0 & 2.6 & \second{7.4}
& 0.0 & 2.6 & \second{5.0} \\

MAI-UI \citep{zhou2025mai}
& 8B
& \second{3.1} & \second{8.8} & \second{10.7}
& \second{3.1} & \second{4.2} & 3.9
& 0.0 & \second{3.7} & 4.2 \\

GUI-Owl \citep{ye2025mobile}
& 32B
& 0.0 & 2.6 & 2.2
& 0.0 & 1.6 & 1.1
& 0.0 & 0.0 & 0.5 \\

UI-Venus-1.5-A3B \citep{team2026ui}
& 30B
& 0.0 & 1.6 & 4.1
& 0.0 & 1.6 & 1.8
& 0.0 & 1.6 & 1.5 \\

\midrule

\rowcolor{gray!12}
\multicolumn{11}{@{}l@{}}{\textbf{\textit{Our model}}} \\

\rowcolor{gray!6}
\textbf{ATMem-UI} (\textcolor{tongyipurple}{Ours})
& \textbf{8B}
& \best{6.2} & \best{11.7} & \best{15.7}
& \best{6.2} & \best{9.4} & \best{9.8}
& \best{3.1} & \best{6.3} & \best{8.2} \\

\bottomrule
\end{tabular}
}
\vspace{-1.2em}
\end{table}

\begin{wraptable}[17]{r}{0.58\columnwidth}
\centering
\caption{
Ablation study on Qwen3-VL-8B-Instruct.
SR denotes success rate. 
Mem. denotes the fraction of tasks in which the agent invokes ATMem at least once during execution.
\besttext{Green}/\secondtext{purple} highlights the best/second-best results.
\textcolor{figcitecolor}{Green $\Delta$ values} indicate SR increases, while \textcolor{purple}{red $\Delta$ values} indicate Mem. reductions.
}
\vspace{-0.5em}
\label{tab:ablation_sft_grpo_strgrpo}

\scriptsize
\setlength{\tabcolsep}{8.5pt}
\renewcommand{\arraystretch}{1.2}

\providecommand{\posdelta}[1]{\textcolor{figcitecolor}{#1}}
\providecommand{\memdelta}[1]{\textcolor{purple}{#1}}

\resizebox{\linewidth}{!}{%
\begin{tabular}{@{}lcccc@{}}
\toprule
\textbf{Training Recipe}
& \multicolumn{2}{c}{\textbf{AndroidWorld}}
& \multicolumn{2}{c}{\textbf{MobileWorld}} \\
\cmidrule(lr){2-3}
\cmidrule(l){4-5}
& \textbf{SR (\%)}
& \textbf{Mem. (\%)}
& \textbf{SR (\%)}
& \textbf{Mem. (\%)} \\
\midrule

\rowcolor{gray!12}
\multicolumn{5}{@{}l@{}}{\textbf{\textit{Qwen3-VL-8B-Instruct}}} \\

Baseline
& 47.6
& --
& 9.4
& -- \\

SFT
& 70.7
& 48.2
& 17.2
& 49.6 \\

\quad $\Delta$ vs. Baseline
& \posdelta{+23.1}
& --
& \posdelta{+7.8}
& -- \\

SFT + GRPO
& \second{74.8}
& \second{45.7}
& \second{20.7}
& \second{44.7} \\

\quad $\Delta$ vs. SFT
& \posdelta{+4.1}
& \memdelta{-2.5}
& \posdelta{+3.5}
& \memdelta{-4.9} \\

\rowcolor{gray!6}
\textbf{SFT + STR-GRPO}
& \best{76.6}
& \best{31.9}
& \best{23.3}
& \best{30.1} \\

\quad $\Delta$ vs. SFT + GRPO
& \posdelta{+1.8}
& \memdelta{-13.8}
& \posdelta{+2.6}
& \memdelta{-14.6} \\

\quad $\Delta$ vs. SFT
& \posdelta{+5.9}
& \memdelta{-16.3}
& \posdelta{+6.1}
& \memdelta{-19.5} \\

\bottomrule
\end{tabular}%
}
\end{wraptable}

As difficulty increases from DC-V1 to DC-V3, the distractor-to-target ratio rises from $1.39\times$ to $3.22\times$, and all agents' performance drops sharply, confirming that denser distractors and tighter data dependencies substantially increase task difficulty.
ATMem-UI-8B degrades more slowly than all end-to-end baselines, maintaining the best end-to-end results at every level.
On DC-V2, it achieves 6.2\% SR, 9.4\% $S_{\mathrm{prog}}$, and 9.8\% $S_{\mathrm{scope}}$, improving over the best non-ours end-to-end result for each metric by 3.1, 5.2, and 2.4 points, respectively.
On DC-V3, it achieves 3.1\% SR, 6.3\% $S_{\mathrm{prog}}$, and 8.2\% $S_{\mathrm{scope}}$, with corresponding gains of 3.1, 2.6, and 3.2 points.
The gains on both $S_{\mathrm{prog}}$ and $S_{\mathrm{scope}}$ show that ATMem improves not only how far the agent progresses, but also how reliably it preserves the correct operation scope under dense distractors.
This pattern is consistent with the role of ATMem's \texttt{Constraints} field and \texttt{skipped} status, which help maintain item-level eligibility information throughout execution and become increasingly important as distractor density grows.

We further include GPT-5.2+UIIns as a framework-level reference built on a strong proprietary planner with explicit external orchestration.
It substantially outperforms all end-to-end agents on DC-V1, achieving 18.8\% SR, 35.4\% $S_{\mathrm{prog}}$, and 38.4\% $S_{\mathrm{scope}}$.
However, its SR also drops from 18.8\% on DC-V1 to 3.1\% on DC-V3, matching ATMem-UI-8B at the highest difficulty level.
This convergence in task success is revealing: under high distractor density, stronger external orchestration alone is not sufficient to avoid failure, suggesting that fine-grained data discrimination and item-level eligibility judgment become critical bottlenecks.
At the same time, the framework retains higher $S_{\mathrm{prog}}$ and $S_{\mathrm{scope}}$, indicating better partial execution quality and pointing to richer intermediate state representation as an important direction for end-to-end agents.
The consistent gains of ATMem-UI across all difficulty levels demonstrate that structured memory is an effective step in this direction, without relying on external orchestration or proprietary planners.

\begin{figure*}[t]
\centering
\includegraphics[width=0.99\textwidth]{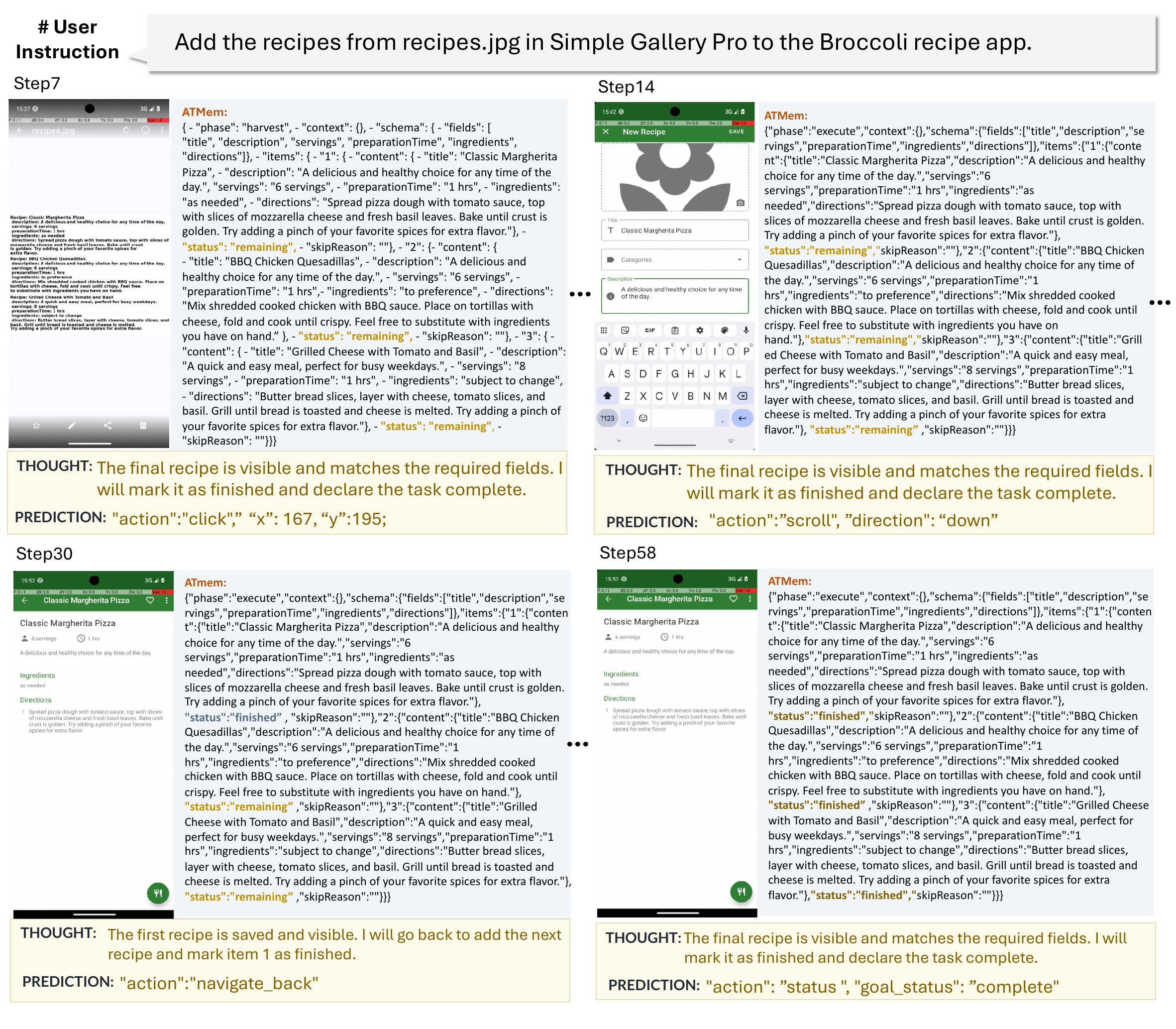}
\vspace{-1.0em}
\caption{
\textbf{Execution case of ATMem-UI on AndroidWorld.}
The figure shows how our agent collects task-relevant data, maintains their structured execution states, and uses these states to guide subsequent actions. 
By tracking which data items are pending or completed, the agent reliably progresses from data collection to task execution and completes the long-horizon workflow.
}
\label{fig:android_world_case}
\end{figure*}

\subsection{Ablation Study}

We conduct an ablation study on Qwen3-VL-8B-Instruct to isolate the contribution of each training stage.
In addition to SR, we report Mem., defined as the fraction of tasks in which the agent invokes ATMem at least once during execution, to measure how selectively the model applies structured memory across different tasks.

\begin{figure*}[t]
\centering
\includegraphics[width=0.99\textwidth]{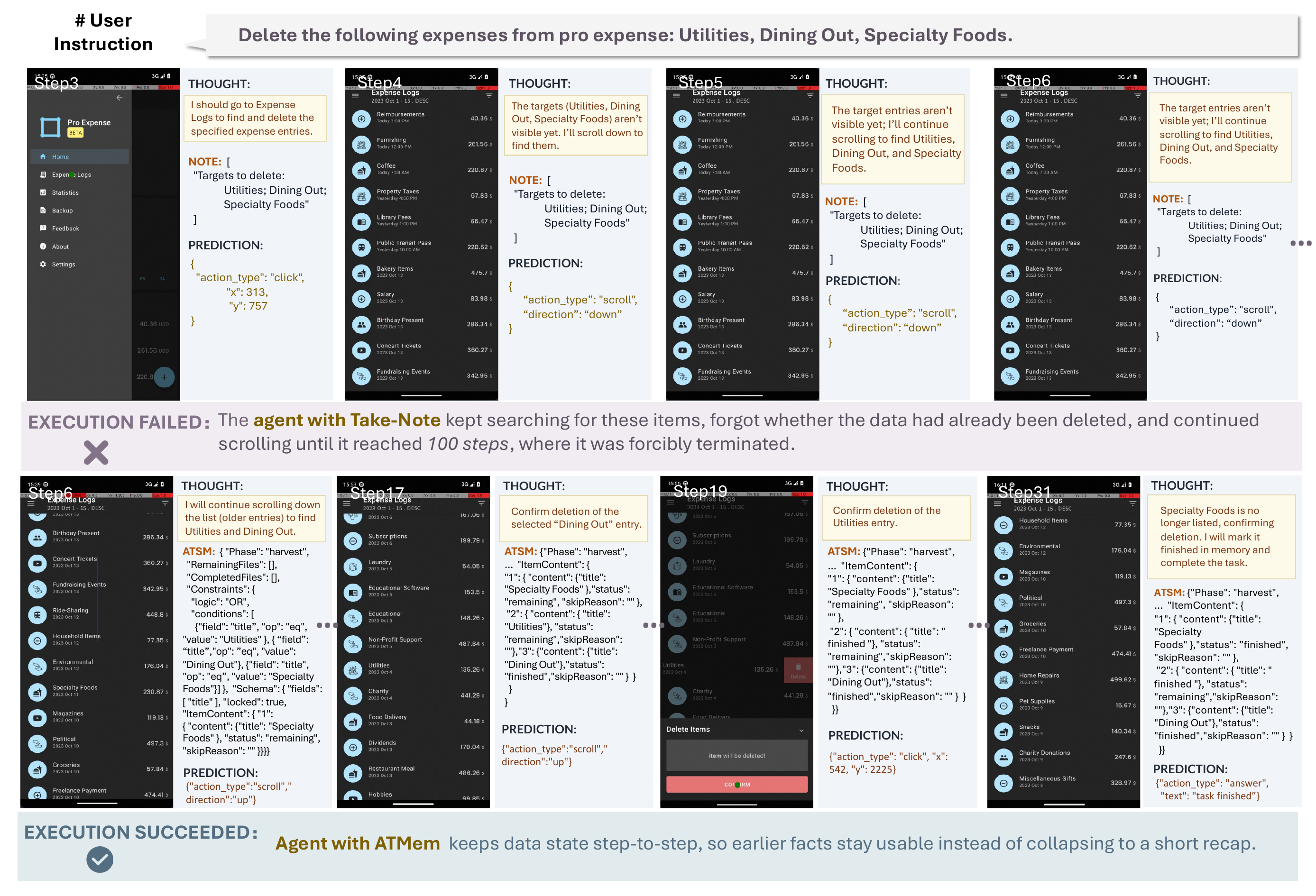}
\caption{
\textbf{Execution trajectory comparison between recording-centric memory and ATMem.}
The flat-memory agent records task information as unstructured notes, but fails to explicitly track which data items have been completed or still require action, leading to repeated search and stuck execution. 
In contrast, the ATMem-based agent maintains structured task data and item-level execution status, enabling more stable progress across complex operations. 
Due to space limits, we omit part of the memory content in the figure. 
}
\label{fig:flat_vs_ASTM}
\end{figure*}

As shown in Table~\ref{tab:ablation_sft_grpo_strgrpo}, the baseline achieves 47.6\% and 9.4\% SR on AndroidWorld and MobileWorld, respectively.
After SFT, SR increases substantially by 23.1 and 7.8 points on the two benchmarks, with Mem. reaching 48.2\% and 49.6\%.
To encourage more selective and effective memory invocation, we introduce STR-GRPO, which estimates the marginal utility of ATMem through paired memory-ON and memory-OFF rollouts.

As Table~\ref{tab:ablation_sft_grpo_strgrpo} suggests, Standard GRPO further improves SR by 4.1 and 3.5 points, and modestly reduces Mem. by 2.5 and 4.9 points.
The reduction is real but limited. 
Through online interaction, the agent receives indirect feedback that some memory-active trajectories do not perform better than simpler ones, which weakly discourages unnecessary invocations.
However, because task-level reward does not distinguish whether a successful trajectory succeeded due to ATMem or despite it, the agent cannot reliably identify which memory invocations are causally useful and which are redundant.
As a result, Mem. remains elevated and the agent continues to apply the ATMem structure to tasks even without extra benefit.

STR-GRPO addresses this issue by estimating the marginal utility of ATMem through paired memory-ON and memory-OFF rollouts within each task group.
This design provides an explicit signal for whether structured memory improves the verifier outcome, allowing the agent to suppress memory updates that do not contribute to success.
Compared with standard GRPO, STR-GRPO reduces Mem. by 13.8 and 14.6 points while further improving SR by 1.8 and 2.6 points.
This joint improvement indicates that many SFT-induced ATMem updates are redundant, and that STR-GRPO removes unnecessary invocations while preserving memory use where it is beneficial.
Relative to SFT, STR-GRPO reduces Mem. by 16.3 and 19.5 points while improving SR by 5.9 and 6.1 points.
These results show that STR-GRPO does not simply discourage memory use, but recalibrates ATMem from a frequently imitated behavior into a more selective, utility-driven mechanism.

\begin{figure*}[t]
\centering
\includegraphics[width=0.99\textwidth]{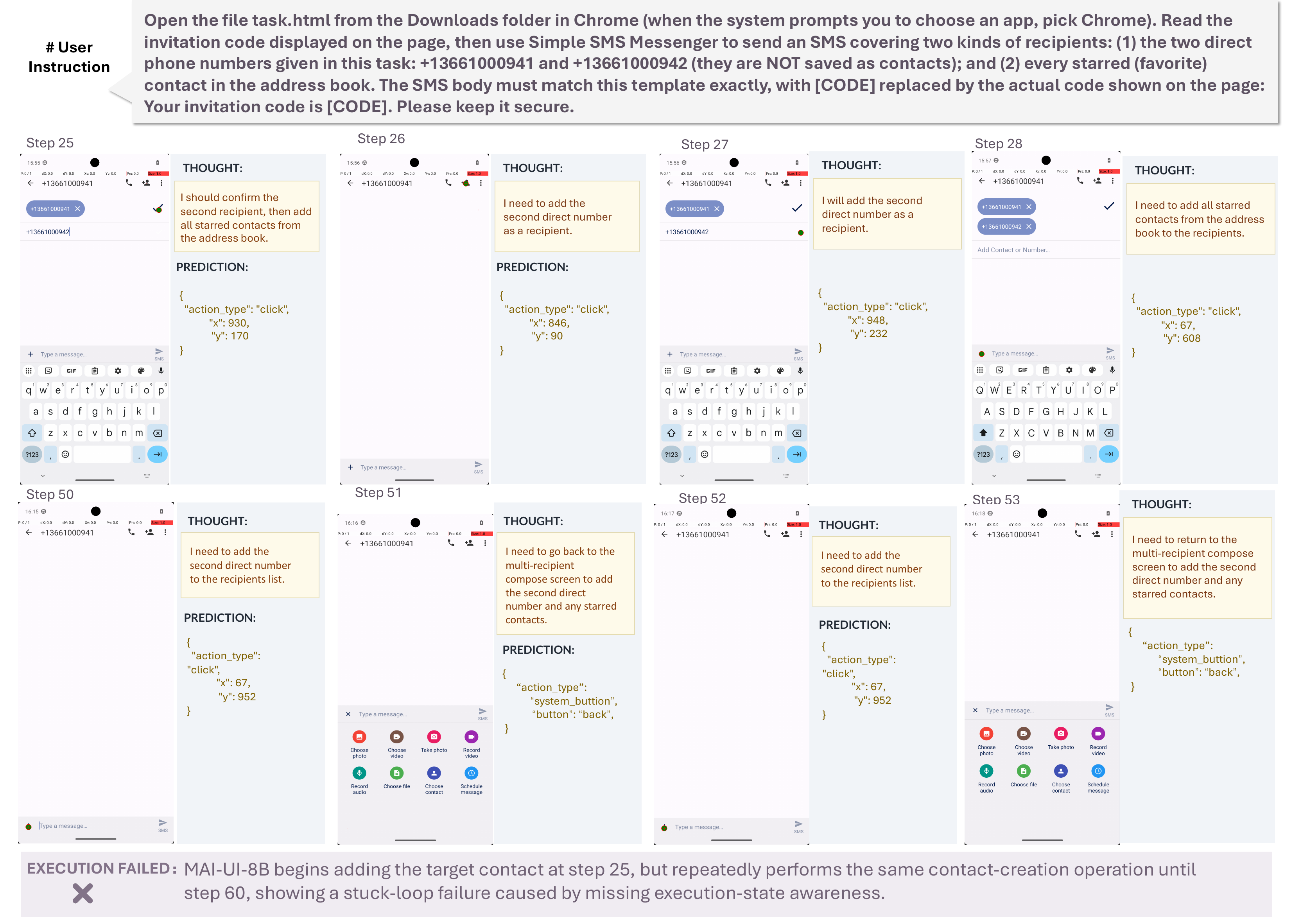}
\caption{
\textbf{Representative failure case on our data-scope benchmark.}
On a data-scope workflow from our benchmark, MAI-UI-8B identifies the target contact information and begins adding the contact at step 25, but then repeats the same contact-creation operation until step 60. 
This stuck-loop behavior suggests that the agent can recover relevant values, but fails to track whether the data operation has already been completed. 
}
\label{fig:mai_ui_8b_failure}
\end{figure*}

\subsection{Qualitative Analysis}
\label{app:qualitative_analysis}

\noindent\textbf{Flat Notes vs.\ ATMem.} To isolate the effect of memory structure, we run the same agentic framework on identical tasks, changing only the memory format.
One setting uses recording-centric memory\citep{wang2025mobile}, while the other uses ATMem.
Figure \ref {fig:flat_vs_ASTM} shows a representative task where the agent must delete three specific expense entries from a list containing many same-schema distractors.
The recording-centric agent stores the target names as a static list and starts searching for them.
However, this memory format does not track which entries have already been deleted.
After removing some targets, the agent revisits previously processed regions and eventually reaches the 100-step limit.
This is a state-tracking failure.
Without item-level completion status, the agent cannot reliably determine which targets remain.

ATMem changes the execution pattern by maintaining explicit status for each target item.
The agent instantiates the three expense entries under \texttt{Schema.ItemContent}, assigns each \texttt{status:remaining}, and encodes the target condition in the \texttt{Constraints} field.
After each deletion, the corresponding item is updated to \texttt{status:finished}, so later decisions are conditioned on the remaining unfinished items.
The task is completed in 31 steps.
This comparison suggests that structured status tracking converts open-ended search into a bounded checklist, reducing the ambiguity that causes flat-note agents to loop.

\noindent\textbf{Representative Failure on DataScope.}
Figure~\ref{fig:mai_ui_8b_failure} shows a representative failure on DataScope.
The task requires sending an SMS to two direct phone numbers and all starred contacts, with the message content derived from a local HTML file.
MAI-UI-8B correctly adds the first recipient and produces a coherent intermediate thought.
However, it then repeatedly attempts to add the second direct number, cycling between the recipient field, the back button, and the compose screen.
By Step 60, the second number is still not confirmed, and the starred contacts have not been processed.
This failure arises from missing execution-state awareness.
The agent has no persistent record that the first recipient has already been added and confirmed.
Because the visible UI state is ambiguous, the agent repeatedly re-evaluates the same situation from interaction history alone and fails to advance to the next subtask.
This pattern is especially common in DataScope, where multi-recipient and multi-item workflows require tracking which specific data units have already been processed.

\noindent\textbf{ATMem in a Long-Horizon Trajectory.}
Figure~\ref{fig:android_world_case} traces ATMem across a successful 58-step trajectory where the agent transfers three recipes from Simple Gallery Pro to the Broccoli recipe app.
During the harvest phase, the agent extracts all recipe fields and stores them as structured entries under \texttt{Schema.ItemContent}, each initialized with \texttt{status:remaining}.
When execution begins, the stored content remains available while the phase field indicates that the agent should start entering recipes into the target app.
As each recipe is saved, ATMem updates the corresponding item to \texttt{status:finished}.
This gives the agent an explicit record of completed and pending items, allowing it to move to the next recipe without re-reading previous screenshots or revisiting completed entries.
By the end of the trajectory, all three recipes are marked finished, and the agent terminates correctly.
This trajectory illustrates three roles of ATMem in long-horizon execution.
During harvest, it stores structured task data that would otherwise need to be recovered from screenshots.
During execution, item status provides a checklist for tracking progress.
Across item transitions, the memory state helps preserve the correct operation scope, enabling the agent to complete a long multi-item workflow without losing track of pending actions.

\subsection{Further Analysis}
\begin{figure*}[t]
\centering
\includegraphics[width=1.0\textwidth]{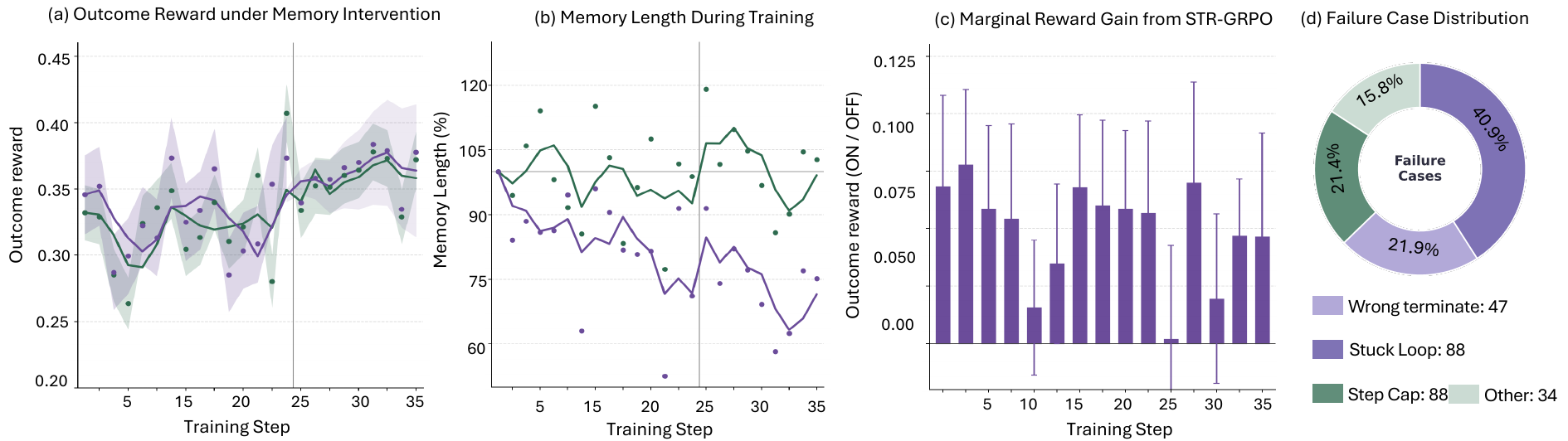}
\caption{
Further analysis of STR-GRPO training dynamics and DataScope failure cases.
\textbf{(a)} Outcome reward under memory-ON and memory-OFF interventions during training, with shaded regions showing rollout-group variance.
Memory-ON consistently achieves higher reward, indicating a sustained benefit from ATMem.
\textbf{(b)} Normalized memory length relative to training step 1.
STR-GRPO substantially reduces memory length, while standard GRPO remains near its initial level.
\textbf{(c)} Marginal reward gain from ATMem, measured as the outcome-reward difference between memory-ON and memory-OFF cohorts within STR-GRPO.
Positive gaps indicate that retained memory invocations provide a measurable benefit.
\textbf{(d)} Failure type distribution on DataScope for MAI-UI-8B, the strongest end-to-end baseline.
Stuck loop is the dominant failure mode at 40.9\%, followed by wrong termination (21.9\%), step cap (21.4\%), and other errors (15.8\%).
}
\label{fig:further_analysis}
\end{figure*}

\noindent\textbf{Training Dynamics under Memory Intervention.}
Figures~\ref{fig:further_analysis} (a)--(c) show how ATMem usage evolves during STR-GRPO training and why selective memory invocation emerges.
Figure~\ref{fig:further_analysis} (a) reports outcome rewards under memory-ON and memory-OFF conditions throughout training.
The memory-ON curve remains consistently higher than the memory-OFF curve, indicating that ATMem provides a positive contribution when it is retained.
Importantly, this gap does not disappear as training progresses.
Even after the model learns to invoke memory more selectively, the remaining memory-active cases still achieve higher reward under the memory-ON condition.
This suggests that STR-GRPO does not improve performance by simply avoiding memory.
Instead, it suppresses low-utility memory updates while preserving cases where structured memory remains beneficial.

Figure \ref{fig:further_analysis} (b) compares memory length under STR-GRPO and standard GRPO.
Under standard GRPO, memory length remains close to its initial level, suggesting that task-level reward alone provides little pressure to reduce memory content.
In contrast, STR-GRPO produces a substantial and sustained reduction in memory length.
Read together with Figure \ref{fig:further_analysis} (a), this reduction is not accompanied by lower outcome reward.
The results suggest that STR-GRPO mainly removes memory updates that provide limited marginal benefit, rather than compressing away useful structured state.

Figure~\ref{fig:further_analysis} (c) directly measures this effect through the ON-minus-OFF reward gap within each task group.
The marginal reward gain from ATMem stays positive across most of the training.
It is larger in early training, when memory is used more broadly, and stabilizes at a smaller but sustained level after redundant memory calls are pruned.
This pattern is consistent with selective invocation.
As low-utility memory updates are removed, the remaining ON-OFF gap reflects the utility of ATMem on tasks that still benefit from explicit state tracking.

\noindent\textbf{Failure Mode Analysis on DataScope.}
Figure~\ref{fig:further_analysis} (d) breaks down the failure cases of MAI-UI-8B, the strongest end-to-end baseline.
Stuck loops are the dominant failure mode at 40.9\%, followed by wrong termination (21.9\%), step cap (21.4\%), and miscellaneous errors, including grounding failures and environment edge cases (15.8\%).

The dominance of stuck loops provides a clear diagnosis of why current end-to-end agents struggle on data-scope tasks.
Stuck loops occur when the agent cannot determine whether a target data item has already been processed, causing it to repeat the same action or revisit completed sub-tasks.
The qualitative example in Figure~\ref{fig:mai_ui_8b_failure} illustrates this pattern, and the aggregate statistics show that it is a recurring failure mode rather than an isolated case.
This is precisely the failure that ATMem is designed to address.
By maintaining an explicit execution status for each item, ATMem allows the agent to identify which items remain pending and which have been completed, without relying only on implicit inference from interaction history.
Wrong termination and step cap failures reflect different remaining challenges.
Wrong termination occurs when the agent declares completion before all required data operations are verified, pointing to limitations in completion checking.
Step cap failures occur when the agent remains in a valid but unresolved execution state and fails to converge within the allowed horizon, pointing to limitations in long-horizon planning and recovery.
These failures are not fully solved by item-level status tracking alone.
They therefore represent the primary remaining challenges for end-to-end agents on data-scope workflows beyond what ATMem directly addresses.

\section{Conclusion}
\label{sec:conclusion}

We presented ATMem, an active task-driving memory framework for long-horizon mobile GUI agents.
Our central argument is that memory in mobile execution should not be treated as a passive archive of past observations.
Instead, task-relevant data should be maintained as an evolving execution state that records what has been collected, what remains pending, which constraints apply, and which operations have already been completed.
ATMem operationalizes this view by organizing task data into structured units with ownership, role, constraints, and status, allowing the agent to condition its next action on current workflow progress rather than relying only on raw interaction history.
To make this memory behavior efficient, we introduced STR-GRPO.
By comparing memory-on/off rollouts and incorporating memory cost into online RL, STR-GRPO teaches the agent not only how to construct ATMem, but also when ATMem is actually useful.
This turns memory invocation from an always-active behavior learned through imitation into a utility-driven policy decision.
We also introduced DataScope, a long-horizon online benchmark that stresses cross-page and cross-application workflows involving data collection, verification, transfer, update, and reuse.
By requiring agents to cover all instruction-relevant targets while filtering same-schema distractors, DataScope exposes failure modes that are difficult to observe from terminal success rate alone.
Across AndroidWorld, MobileWorld, and DataScope, ATMem-UI improves long-horizon execution while reducing unnecessary memory usage.
Our failure analysis further shows that stuck loops, premature termination, and incomplete operation tracking remain key bottlenecks.

\appendix
\section{Appendix}

\subsection{Online RL Training Environment and Infrastructure}
\label{app:online_rl_infra}

Long-horizon mobile-agent training requires rollouts to be collected through real multi-turn interactions with the environment. 
This makes a standard synchronous RL pipeline inefficient, since policy updates must wait for many long and heterogeneous trajectories to finish. 
The trajectories themselves are also expensive to optimize over: each rollout may concatenate multiple GUI observations, actions, reasoning traces, and ATMem states, resulting in extremely long multimodal contexts that exceed single-GPU memory limits. 
Following MAI-UI~\citep{zhou2025mai}, we build an asynchronous on-policy RL framework based on \texttt{verl} and HybridFlow~\citep{sheng2024hybridflow} to support scalable online training.

\noindent\textbf{Asynchronous rollout collection.}
We implement a customized agent loop that asynchronously dispatches rollout requests to a pool of inference servers hosting the latest policy model. 
This design reduces GPU idle time caused by long multi-turn environment interactions and improves rollout throughput. 
The agent loop also supports asynchronous environment interaction and session management, including backup sessions for failure recovery and replacement. 
On the inference side, request load balancing and prefill caching are used to accelerate generation under long interaction histories. 
Although rollouts are collected asynchronously, training remains on-policy: each rollout batch is generated by the current policy version and stored with the corresponding behavior log probabilities for policy optimization.

\noindent\textbf{Hybrid parallelism for long trajectories.}
To optimize policies over ultra-long trajectories, we adopt Megatron-style hybrid parallelism, including tensor parallelism, pipeline parallelism, and context parallelism. 
This partitions both model computation and long sequence contexts across multiple GPUs, allowing end-to-end policy updates while keeping per-GPU memory usage manageable. 
We additionally downsample GUI screenshots to half resolution, which reduces visual-token length and training cost without degrading empirical performance.

Overall, this infrastructure enables stable online RL for long-horizon mobile agents by decoupling rollout collection from policy optimization, improving inference throughput, and supporting training over extremely long multimodal trajectories.

\subsection{Training Infrastructure and Hyperparameters}
\label{app:training_infra}

We provide the training infrastructure and main hyperparameters in Table~\ref{tab:training_infra}. 
For SFT, we train on verified rollout trajectories using 128 NVIDIA H20 GPUs across 16 nodes. 
Each node contains 8 GPUs and a 96-core CPU with approximately 900GB of memory. 
The SFT stage takes about 2 hours. 
We train the language model while freezing the vision encoder, using bfloat16 precision, FlashAttention, gradient checkpointing, Liger kernel acceleration, and DeepSpeed ZeRO-2 offload.

For online RL, we use the same 128-GPU training configuration and train for about 3 days. 
The mobile environments are deployed on bare-metal servers. 
In our main run, we use 128 active environment containers for rollout collection and maintain additional backup containers for failure recovery. 
Each environment server has 96 CPU cores and 384GB of memory. 
The rollout system sets the maximum response length to 81,920 tokens. 
The RL actor is initialized from the SFT checkpoint.

The RL system is built on a \texttt{verl}-based multi-turn agent training stack. 
For generation and policy optimization, we use tensor parallelism with degree 2, context parallelism with degree 2, and pipeline parallelism with degree 1. 
The rollout engine uses a GPU memory utilization ratio of 0.55, disables replay, and performs checkpointing and evaluation every epoch.

\subsection{ATMem-Augmented Agent Execution Prompt}
\label{app:agent_execution_prompt}

During evaluation, our agent uses the structured execution prompt in Box~\ref{box:agent_execution_prompt}. 
It constrains each response to include reasoning, a single executable tool call, and an optional ATMem memory update.

\begin{tcolorbox}[
title=ATMem-Augmented Agent Execution Prompt,
breakable,
fonttitle=\scriptsize\bfseries\color{white},
fontupper=\scriptsize,
colback=gray!3,
colframe=gray!35,
colbacktitle=gray!300,
coltitle=white,
arc=0.8mm,
boxsep=1mm,
left=1mm,
right=1mm,
top=1mm,
bottom=1mm
]
\label{box:agent_execution_prompt}

\textbf{\#\# Task.}
You are a GUI agent. You are given a task goal, previous action history, and screenshots of the current screen. 
You need to perform the \textbf{NEXT} action to complete the task.

\medskip
\textbf{\#\# Output Format.}
For each step, output exactly \textbf{THREE} blocks in this order:

\begin{tcolorbox}[
colback=black!3,
colframe=black!10,
arc=0.6mm,
boxsep=0.6mm,
left=0.8mm,
right=0.8mm,
top=0.6mm,
bottom=0.6mm
]
{\tiny\ttfamily
<thinking>\\
... return the thinking process\\
</thinking>\\
<tool\_call>\\
\{"name": "mobile\_use", "arguments": <args-json-object>\}\\
</tool\_call>\\
<memory>\\
\{ ...valid JSON dict or \{\} ... \}\\
</memory>
}
\end{tcolorbox}

\textbf{Rules.}
\begin{itemize}[leftmargin=2.8mm,itemsep=0.2ex,topsep=0.2ex]
  \item Output exactly \textbf{one} tool call per step.
\end{itemize}

\medskip
\textbf{\#\# Action Space.}
Choose exactly \textbf{one} action per step:

{\scriptsize
\renewcommand{\arraystretch}{1.08}
\begin{tabularx}{\linewidth}{@{}l >{\raggedright\arraybackslash}X@{}}
\toprule
\textbf{Action} & \textbf{JSON Format} \\
\midrule
\texttt{open\_app} 
& {\tiny\ttfamily \{"action": "open\_app", "app\_name": "<Available Apps>"\}} \\

\texttt{click} 
& {\tiny\ttfamily \{"action": "click", "coordinate": [x, y]\}} \\

\texttt{long\_press} 
& {\tiny\ttfamily \{"action": "long\_press", "coordinate": [x, y]\}} \\

\texttt{input\_text} 
& {\tiny\ttfamily \{"action": "input\_text", "text": ""\}} \\

\texttt{scroll} 
& {\tiny\ttfamily \{"action": "scroll", "direction": "up|down|left|right"\}} \\

\texttt{drag} 
& {\tiny\ttfamily \{"action": "drag", "coordinate": [x, y], "direction": "up|down|left|right"\}} \\

\texttt{navigate\_back} 
& {\tiny\ttfamily \{"action": "navigate\_back"\}} \\

\texttt{navigate\_home} 
& {\tiny\ttfamily \{"action": "navigate\_home"\}} \\

\texttt{wait} 
& {\tiny\ttfamily \{"action": "wait"\}} \\

\texttt{status} 
& {\tiny\ttfamily \{"action": "status", "goal\_status": "complete|infeasible"\}} \\

\texttt{answer} 
& {\tiny\ttfamily \{"action": "answer", "text": ""\}} \\
\bottomrule
\end{tabularx}
}

\medskip
\textbf{\#\# Hard Constraints.}
\begin{itemize}[leftmargin=2.8mm,itemsep=0.2ex,topsep=0.2ex]
  \item Use \texttt{answer} only when the task can be completed without interacting with the device UI.
  \item Otherwise, use device actions and do not answer directly.
  \item Output \texttt{goal\_status="complete"} only after verifying the result on screen.
\end{itemize}

\medskip
\textbf{\#\# When to Use Memory.}
Use \texttt{<memory>} only when cross-step storage or comparison is needed; otherwise output \texttt{\{\}}.

\medskip
\textbf{\#\# Memory Schema.}
When used, memory must follow this structure:

\begin{tcolorbox}[
colback=black!3,
colframe=black!10,
arc=0.6mm,
boxsep=0.6mm,
left=0.8mm,
right=0.8mm,
top=0.6mm,
bottom=0.6mm
]
{\tiny\ttfamily
\{\\
\ \ "phase": "harvest" | "execute",\\
\ \ "context": \{\},\\
\ \ "schema": \{\\
\ \ \ \ "fields": []\\
\ \ \},\\
\ \ "items": \{\\
\ \ \ \ "1": \{\\
\ \ \ \ \ \ "content": \{\},\\
\ \ \ \ \ \ "status": "remaining" | "finished" | "skipped",\\
\ \ \ \ \ \ "skipReason": ""\\
\ \ \ \ \}\\
\ \ \}\\
\}
}
\end{tcolorbox}

\textbf{Rules.}
\begin{itemize}[leftmargin=2.8mm,itemsep=0.2ex,topsep=0.2ex]
  \item Store only structured values in memory; do not store raw text dumps.
\end{itemize}

\medskip
\textbf{\#\# Memory Usage Protocol.}
\begin{itemize}[leftmargin=2.8mm,itemsep=0.2ex,topsep=0.2ex]
  \item If using memory, start with \texttt{phase="harvest"} and switch to \texttt{phase="execute"} once enough information is collected.
  \item Update an item status to \texttt{finished} in the same step after a successful operation.
\end{itemize}

\end{tcolorbox}

\begin{table}[t]
\centering
\caption{Training infrastructure and main hyperparameters.}
\label{tab:training_infra}
\small
\setlength{\tabcolsep}{5pt}
\renewcommand{\arraystretch}{1.12}
\begin{tabular}{lcc}
\toprule
\textbf{Configuration} & \textbf{SFT} & \textbf{Online RL} \\
\midrule
Base model & Qwen3-VL-4B/8B-Instruct & SFT checkpoint \\
Training GPUs & 128 NVIDIA H20 & 128 NVIDIA H20 \\
Node setup & 16 nodes, 8 GPUs/node & 16 nodes, 8 GPUs/node \\
CPU / memory per training node & 96 cores / $\sim$900GB & 96 cores / $\sim$900GB \\
Training time & $\sim$2 hours & $\sim$3 days \\
Precision & bfloat16 & bfloat16 \\
Max sequence / response length & 20,768 & 81,920 \\
Vision encoder & Frozen & -- \\
Parallelism & ZeRO-2 offload & TP=2, PP=1, CP=2 \\
Batch size & 1 per GPU & global 32, mini-batch 32 \\
Rollout group size & -- & 16 \\
Screenshot resolution & max pixels 12,845,056 & $1600\times720$ \\
Environment servers & -- & 3 bare-metal servers \\
Environment server config & -- & 96 cores / 384GB each \\
Environment containers & -- & 200 total, 128 active \\
\bottomrule
\end{tabular}
\end{table}

\subsection{Data-Centric Benchmark Details}
\label{app:benchmark_details}

This section provides detailed statistics of our Data-Centric benchmark. 
Each task family is designed as a data-dependent mobile workflow involving one or more applications, where the agent must identify target data, avoid confusable distractors, and perform the required data operations. 
Here, confusable distractors refer to non-target entries that share the same schema or data type with target entries and are generated with template-level similarity constraints. 
To construct controlled difficulty levels, we instantiate the same task families across DC-V1, DC-V2, and DC-V3 while progressively increasing the number of target entries, confusable distractors, and required operations. 
This design keeps the underlying app workflow largely comparable across subsets, while scaling difficulty through data volume and distractor density.

Table~\ref{tab:benchmark_family_details_color} reports task-family-level details. 
For each subset, \textbf{T/N/O} denotes the number of target entries, confusable distractors, and required operations, respectively. 
The application column lists the apps involved in each workflow using color-coded labels. 
A ``--'' indicates that the corresponding task variant is not included in that subset after filtering unreachable or invalid instances.

\begingroup
\footnotesize
\setlength{\tabcolsep}{4pt}
\renewcommand{\arraystretch}{1.12}
\rowcolors{3}{gray!6}{white}
\vspace{2.0em}
\begin{longtable}{p{0.30\linewidth}p{0.18\linewidth}c>{\centering\arraybackslash}p{0.10\linewidth}>{\centering\arraybackslash}p{0.10\linewidth}>{\centering\arraybackslash}p{0.10\linewidth}}

\caption{
\textbf{Task-family details across benchmark subsets.}
Each subset cell reports \textbf{T/N/O}, where \textbf{T}, \textbf{N}, and \textbf{O} denote the numbers of target entries, same-semantics distractors, and required operations, respectively. 
Applications are shown as color-coded labels for readability.
}
\label{tab:benchmark_family_details_color}\\

\toprule
\textbf{Task Family} & \textbf{Apps} & \textbf{\#Apps} & \textbf{DC-V1} & \textbf{DC-V2} & \textbf{DC-V3} \\
\midrule
\endfirsthead

\toprule
\textbf{Task Family} & \textbf{Apps} & \textbf{\#Apps} & \textbf{DC-V1} & \textbf{DC-V2} & \textbf{DC-V3} \\
\midrule
\endhead

\midrule
\multicolumn{6}{r}{\textit{Continued on next page}} \\
\endfoot

\bottomrule
\endlastfoot

CalendarFocusTracksRetroPlaylist 
& \makecell[l]{\appCalendar\\[2pt]\appRetroMusic} 
& 2 & 0/4/3 & 0/13/3 & 0/20/3 \\

CalendarMeetingPrepAndNotify 
& \makecell[l]{\appCalendar} 
& 1 & 5/13/5 & 20/43/20 & 36/100/36 \\

CalendarTodayAlarmsInClock 
& \makecell[l]{\appCalendar\\[2pt]\appClock} 
& 2 & 2/7/2 & 3/10/3 & 4/17/4 \\

CallLogContactTodoMarkorSync 
& \makecell[l]{\appContacts\\[2pt]\appMarkor\\[2pt]\appPhone\\[2pt]\appTasks} 
& 4 & 2/0/2 & 3/6/3 & 3/7/3 \\

CallLogTopContactsFavorite 
& \makecell[l]{\appContacts\\[2pt]\appPhone} 
& 2 & 13/1/13 & 18/14/18 & 22/29/22 \\

ChromeInviteCodeAndSmsSend 
& \makecell[l]{\appChrome\\[2pt]\appMessages} 
& 2 & 3/4/3 & 4/7/4 & 9/18/9 \\

ExpensePurgeHousingFoodLogMarkor 
& \makecell[l]{\appMarkor\\[2pt]\appExpense} 
& 2 & 0/0/3 & 0/0/3 & 0/0/5 \\

FilesClosedProjectContactDelete 
& \makecell[l]{\appContacts\\[2pt]\appFiles} 
& 2 & 2/4/2 & 3/12/3 & 3/35/3 \\

FilesDatedReimburseTxtToExpense 
& \makecell[l]{\appFiles\\[2pt]\appExpense} 
& 2 & 3/0/3 & 3/0/3 & 4/0/4 \\

FilesPartyMenuRecipesToBroccoli 
& \makecell[l]{\appBroccoli\\[2pt]\appFiles} 
& 2 & 2/0/2 & 2/0/2 & 2/0/2 \\

FilesSetlistExportRetroM3u 
& \makecell[l]{\appFiles\\[2pt]\appRetroMusic} 
& 2 & 0/3/4 & 0/6/4 & 0/9/4 \\

FilesWorkEventsToCalendar 
& \makecell[l]{\appCalendar\\[2pt]\appFiles} 
& 2 & 3/6/3 & 6/10/6 & 8/17/8 \\

FilesWorkMergeToMarkorCsv 
& \makecell[l]{\appFiles\\[2pt]\appMarkor} 
& 2 & 3/3/3 & 4/3/4 & 6/8/6 \\

GalleryImageListToMarkor 
& \makecell[l]{\appGallery\\[2pt]\appMarkor} 
& 2 & 4/1/4 & 4/6/4 & 6/12/6 \\

GalleryVlcArtistPlaylistsFromImages 
& \makecell[l]{\appGallery\\[2pt]\appVLC} 
& 2 & 2/3/2 & 3/12/3 & 4/21/4 \\

MarkorTaskAssignmentDistribute 
& \makecell[l]{\appMarkor} 
& 1 & 2/5/2 & 4/25/4 & 5/60/5 \\

RecipeNutFreePurgeLogMarkor 
& \makecell[l]{\appBroccoli\\[2pt]\appMarkor} 
& 2 & 6/3/6 & 7/3/7 & 9/3/9 \\

RetroLightMusicPartyMarkorSms 
& \makecell[l]{\appMarkor\\[2pt]\appMessages\\[2pt]\appRetroMusic} 
& 3 & 2/2/2 & 2/6/2 & 3/8/3 \\

SmsAgentPhoneToMarkorTxt 
& \makecell[l]{\appMarkor\\[2pt]\appMessages} 
& 2 & 4/10/4 & 4/7/4 & 6/30/6 \\

SmsAlbumQueueRetroMarkor 
& \makecell[l]{\appMarkor\\[2pt]\appMessages\\[2pt]\appRetroMusic} 
& 3 & 2/8/2 & 2/13/2 & 2/21/2 \\

SmsColleagueTripReimburseToExpense 
& \makecell[l]{\appMessages\\[2pt]\appExpense} 
& 2 & 1/4/1 & 1/4/1 & 2/15/10 \\

SmsDavidBugfixToCalendar 
& \makecell[l]{\appCalendar\\[2pt]\appMessages} 
& 2 & 3/6/3 & 4/8/4 & 6/10/6 \\

SmsDavidClientsToContacts 
& \makecell[l]{\appContacts\\[2pt]\appMessages} 
& 2 & 3/7/3 & 3/18/3 & 4/48/4 \\

SmsDavidTodoToTasks 
& \makecell[l]{\appMessages\\[2pt]\appTasks} 
& 2 & 1/3/1 & 2/10/2 & 3/25/3 \\

SmsExpenseAuditAndReport 
& \makecell[l]{\appMessages\\[2pt]\appExpense} 
& 2 & 0/2/2 & 0/6/6 & 0/15/12 \\

SmsFamilyRecipeToBroccoli 
& \makecell[l]{\appBroccoli\\[2pt]\appMessages} 
& 2 & 2/4/2 & 3/9/3 & 4/13/4 \\

SmsFriendRecommendationsVlcPlaylists 
& \makecell[l]{\appMessages\\[2pt]\appVLC} 
& 2 & 2/3/2 & 5/13/5 & 7/20/7 \\

SmsPotluckRecipeBroccoliCalendar 
& \makecell[l]{\appBroccoli\\[2pt]\appCalendar\\[2pt]\appMessages} 
& 3 & 4/6/4 & 4/6/4 & 5/9/5 \\

SmsStockDataToMarkorCsv 
& \makecell[l]{\appMarkor\\[2pt]\appMessages} 
& 2 & 2/5/2 & 4/12/4 & 7/27/7 \\

SmsWorkPlanToMarkorTodo 
& \makecell[l]{\appMarkor\\[2pt]\appMessages\\[2pt]\appTasks} 
& 3 & 1/3/1 & 3/10/3 & 5/25/5 \\

SmsWorkoutPlaylistRetroMarkor 
& \makecell[l]{\appMarkor\\[2pt]\appMessages\\[2pt]\appRetroMusic} 
& 3 & 1/3/1 & 1/9/1 & 1/20/1 \\

VlcPlaylistsToMarkorMdFiles 
& \makecell[l]{\appFiles\\[2pt]\appMarkor\\[2pt]\appVLC} 
& 3 & 2/8/2 & 2/11/2 & 2/17/2 \\

\end{longtable}
\endgroup

\subsection{Example Workflow Template of Our Online Benchmark}
\label{app:workflow_template_example}

To illustrate our data-centric workflow specification, Listing~\ref{lst:workflow_template_example} presents a compact example from one of our 32 workflow templates. 
The example is shortened for readability but preserves the main fields used in benchmark construction. 
The \texttt{task}, \texttt{goal}, and \texttt{apps} fields define the task family, user-facing instruction, and involved mobile applications. 
The \texttt{custom\_params\_template} and \texttt{initialization\_example} specify the concrete data injected into the environment, including target files, app-specific records, and confusable distractors. 
The \texttt{expected\_output\_example} field defines the required terminal state after successful execution, while \texttt{\_verification} lists the app-level checks used by the verifier. 
The \texttt{task\_extension\_suggestions} field further describes how the same workflow can be instantiated with different data values and scaled to harder variants, such as adding more target entries or distractors. 
Together, these fields bind the task goal, environment initialization, expected outcome, and automatic verification into a single executable task specification.

\subsection{Environment Initialization and Verifier Calibration}
\label{app:env_init_verifier}

Following AndroidWorld's~\citep{rawles2024androidworld} app-state-based task construction practice, we implement app-specific initialization and verification functions for each workflow template. 
For each involved application, the initializer writes the required target data and confusable distractors into the corresponding app storage, such as SQLite-backed records or file-system entries. 
This ensures that all task-relevant data and distractor data are visible to the agent through normal GUI interaction, rather than being provided only in the instructions.

For each task family, we further implement a task-specific verifier that checks whether the final environment state satisfies the expected outcome. 
The verifier inspects app-level terminal states, such as created files, modified records, sent messages, calendar entries, contact updates, or untouched distractors, and converts them into rule-level scores. 
This enables automatic evaluation of both terminal success and partial app-/data-level progress.

To validate verifier reliability, we perform verifier calibration with controlled terminal-state variants. 
Given the expected result of each task, GPT-5.2 generates multiple perturbed terminal states, including successful completion, missing one or more required data operations, extra operations on non-target data, missing operations in a particular app, and incorrect field values. 
We then initialize the environment directly with each controlled terminal state, treating it as a simulated post-execution state, and run the corresponding verifier. 
A verifier is retained only if its score matches the expected score for all controlled variants; otherwise, the task case is returned for manual correction. 
This calibration process helps ensure that our automatic evaluators correctly distinguish complete success, partial completion, over-operation, and incorrect data manipulation.

\subsection{Prompt Template for Benchmark Task Generation}
\label{app:task_generation_prompt}

To instantiate diverse task variants from each workflow template, we use an LLM-based task generator with a structured system prompt. 
The prompt constrains generation to the same logical task class as the source template: the involved applications, cross-app workflow, output types, and structural requirements must remain unchanged, while concrete data values and scenario entities can vary. 
Difficulty is controlled only through the number of target data entries and confusable distractors, rather than by adding new steps or extra instructions. 
The prompt also requires each generated variant to include initialization data and expected outputs following the source template schema, so that the resulting task can be initialized in the environment and evaluated automatically by the verifier. 
A shortened version of the system prompt is shown in Box~\ref{box:task_generation_prompt}.

\begin{tcolorbox}[
title=System Prompt for Data-Centric Task Variant Generation,
breakable,
fonttitle=\scriptsize\bfseries,
fontupper=\scriptsize,
colback=gray!3,
colframe=gray!35,
colbacktitle=gray!300,
arc=0.8mm,
boxsep=1mm,
left=1mm,
right=1mm,
top=1mm,
bottom=1mm
]
\label{box:task_generation_prompt}
\textbf{\#\# Task.}
Generate feasible GUI automation task variants for an Android long-horizon benchmark.

\medskip
\textbf{\#\# Inputs.}
\begin{itemize}[leftmargin=2.8mm,itemsep=0.2ex,topsep=0.2ex]
  \item \texttt{requested\_num\_tasks}: exact number of variants to output.
  \item \texttt{per\_variant\_difficulty}: ordered list of difficulty labels.
  \item \texttt{full\_benchmark\_template}: complete benchmark JSON for one task family.
  \item Optional trajectory screenshots: UI references only.
\end{itemize}

\medskip
\textbf{\#\# Task-Class Constraints.}
Generate variants for the \textbf{same logical task class} as the source template.
\begin{itemize}[leftmargin=2.8mm,itemsep=0.2ex,topsep=0.2ex]
  \item Keep the same apps, cross-app workflow, and output types.
  \item Scenario wording and concrete values may change, but apps, steps, and structural requirements must not be added or removed.
  \item \texttt{task\_goal} and \texttt{cn\_task\_goal} should remain concise, specific, and agent-actionable.
\end{itemize}

\medskip
\textbf{\#\# Difficulty Model.}
Difficulty is increased only through \textbf{target data volume} and \textbf{confusable distractor density}, not by adding new steps or extra instructions.

{\scriptsize
\renewcommand{\arraystretch}{1.08}
\begin{tabularx}{\linewidth}{@{}l >{\raggedright\arraybackslash}X >{\raggedright\arraybackslash}X@{}}
\toprule
\textbf{Difficulty} & \textbf{Operational / Target Data} & \textbf{Noise / Distractor Data} \\
\midrule
\texttt{easy} 
& 1--2 target sub-groups. 
& 2--5 noise items, clearly distinct from targets. \\

\texttt{medium} 
& 2--3 target sub-groups; roughly 1.5$\times$ easy counts. 
& 5--8 noise items with structural similarity. \\

\texttt{hard} 
& 3--5 target sub-groups; roughly 2$\times$ easy counts. 
& 8--12 highly similar noise items, including near-duplicate names, formats, or partial-match fields. \\

\texttt{very\_hard} 
& 5--8 target sub-groups; roughly 3$\times$ easy counts. 
& 12--20 highly confusable noise items, including trap entries with one missing or close-but-wrong field. \\
\bottomrule
\end{tabularx}
}

\medskip
\textbf{\#\# Reference and Ground Truth.}
When an \texttt{lh\_v1\_reference} is provided, use it as the authoritative reference for:
\begin{enumerate}[leftmargin=3.2mm,itemsep=0.2ex,topsep=0.2ex]
  \item \texttt{task\_goal} and \texttt{cn\_task\_goal} phrasing;
  \item \texttt{expected\_output} structure, including top-level keys, nesting, and field names;
  \item \texttt{init\_data.custom\_params} schema.
\end{enumerate}
Only concrete values, counts, and scenario entities may change.

\medskip
\noindent\textbf{\#\# Expected Output Rules.}
\begin{itemize}[leftmargin=2.8mm,itemsep=0.2ex,topsep=0.2ex]
  \item Fill \texttt{expected\_output} to exactly match what a correct agent should produce for the current variant.
  \item Use the source template's \texttt{expected\_output\_example} as the structural reference.
  \item Do not emit \texttt{rule\_validation}; it is derived automatically from \texttt{expected\_output}.
  \item Do not emit keys starting with \texttt{\_} inside \texttt{expected\_output}.
\end{itemize}

\medskip
\textbf{\#\# Initialization Data.}
\begin{itemize}[leftmargin=2.8mm,itemsep=0.2ex,topsep=0.2ex]
  \item \texttt{init\_data} must not be null.
  \item Follow the structural layout of \texttt{full\_benchmark\_template.initialization\_example}.
  \item Ensure unique names, phone numbers, and IDs for all contacts or records.
  \item Strip all \texttt{\_}-prefixed keys from emitted initialization data.
\end{itemize}

\medskip
\textbf{\#\# Output Format.}
Return a single valid JSON object only. Do not include markdown fences, comments, or extra text.

\begin{tcolorbox}[
colback=black!3,
colframe=black!10,
arc=0.6mm,
boxsep=0.6mm,
left=0.8mm,
right=0.8mm,
top=0.6mm,
bottom=0.6mm
]
{\tiny\ttfamily
\{\\
\ \ "task\_type": "<exact task type from source template>",\\
\ \ "variants": [\\
\ \ \ \ \{\\
\ \ \ \ \ \ "variant\_id": 1,\\
\ \ \ \ \ \ "difficulty": "easy|medium|hard|very\_hard",\\
\ \ \ \ \ \ "task\_goal": "<same task class; values may change>",\\
\ \ \ \ \ \ "cn\_task\_goal": "<Chinese task goal>",\\
\ \ \ \ \ \ "init\_data": \{\\
\ \ \ \ \ \ \ \ "task\_type": "<same as task\_type>",\\
\ \ \ \ \ \ \ \ "custom\_params": \{ "<seeded data fields from template>" \}\\
\ \ \ \ \ \ \},\\
\ \ \ \ \ \ "expected\_output": \{ "<ground truth with reference structure>" \},\\
\ \ \ \ \ \ "rationale": "<=120 words describing target/noise counts>"\\
\ \ \ \ \}\\
\ \ ]\\
\}
}
\end{tcolorbox}

\end{tcolorbox}

\clearpage

\begin{lstlisting}[
style=jsonstyle,
caption={Compact example of a data-centric workflow template. Arrays are shortened for readability while preserving the main template structure.},
label={lst:workflow_template_example}
]
{
  "task": "CalendarMeetingPrepAndNotify",
  "goal": "Check next week's meetings in Simple Calendar Pro. For each target meeting, find related pending tasks in Tasks whose notes mention the meeting title, create a Markor agenda file named {meeting_title}_agenda.md, look up attendee phone numbers in Contacts, and send the agenda via Simple SMS Messenger.",
  "apps": [
    "Simple Calendar Pro",
    "Tasks",
    "Contacts",
    "Markor",
    "Simple SMS Messenger"
  ],
  "task_type": "long-horizon",

  "_why_hard": [
    "Five apps are chained: Calendar -> Tasks -> Markor -> Contacts -> SMS.",
    "Target meetings, tasks, and contacts are mixed with confusable distractors.",
    "The SMS body must match the corresponding meeting agenda."
  ],

  "task_extension_suggestions": {
    "1": "Increase the number of target meetings; each meeting requires an independent agenda file and SMS messages.",
    "2": "Increase noise_events, noise_tasks, and noise_contacts.",
    "3": "Add attendees without matching contacts to test skipping or fault tolerance.",
    "4": "Require strict agenda filename or section formatting."
  },

  "difficulty_control": {
    "easy": {
      "target_meetings": 1,
      "attendees_per_meeting": 2,
      "related_tasks_per_meeting": 2,
      "noise_events": 3,
      "noise_tasks": 5,
      "noise_contacts": 5
    },
    "medium": {
      "target_meetings": "2-3",
      "attendees_per_meeting": "2-4",
      "related_tasks_per_meeting": "2-5",
      "noise_events": 8,
      "noise_tasks": 15,
      "noise_contacts": 20
    },
    "hard": {
      "target_meetings": "3-4",
      "attendees_per_meeting": "3-6",
      "related_tasks_per_meeting": "3-8",
      "noise_events": 15,
      "noise_tasks": 35,
      "noise_contacts": 50
    }
  },

  "custom_params_template": {
    "calendar_events": {
      "target_events": [
        {
          "title": "Vendor Contract Sync",
          "date": "2023-10-24",
          "time": "10:00",
          "duration_mins": 60,
          "location": "Meeting Room 2B",
          "description": "Align on contract renewal timeline and action items.",
          "attendees": ["Nina Patel", "Omar Reyes"]
        }
      ],
      "noise_events": [
        {
          "title": "Team Lunch",
          "date": "2023-10-25",
          "time": "12:00",
          "duration_mins": 60,
          "description": "Lunch at the cafe.",
          "attendees": []
        }
      ]
    },

    "tasks_app_data": {
      "target_tasks": [
        {
          "title": "Draft agenda for vendor sync",
          "completed": 0,
          "deleted": 0,
          "notes": "For Vendor Contract Sync: propose topics and timing."
        },
        {
          "title": "Collect renewal milestones",
          "completed": 0,
          "deleted": 0,
          "notes": "Needed for Vendor Contract Sync."
        }
      ],
      "noise_tasks": [
        {
          "title": "Buy dish soap",
          "completed": 0,
          "deleted": 0,
          "notes": null
        }
      ]
    },

    "contacts_data": {
      "target_contacts": [
        {
          "name": "Nina Patel",
          "number": "+13661100111"
        },
        {
          "name": "Omar Reyes",
          "number": "+13661100112"
        }
      ],
      "noise_contacts": [
        {
          "name": "Mia Chen",
          "number": "+13661100202"
        }
      ]
    },

    "output_config": {
      "markor_file_template": "{meeting_title}_agenda.md",
      "markor_content_format": "# {meeting_title}\n- Date: {date}\n- Time: {time}\n- Location: {location}\n- Attendees: {attendees}\n\n## Related Tasks\n{task_list}",
      "sms_format": "Agenda for {meeting_title} ({date} {time}): {task_list}. Location: {location}."
    }
  },

  "initialization_example": {
    "task_type": "CalendarMeetingPrepAndNotify",
    "task_idx": 0,
    "custom_params": {
      "calendar_events": "instantiated from custom_params_template.calendar_events",
      "tasks_app_data": "instantiated from custom_params_template.tasks_app_data",
      "contacts_data": "instantiated from custom_params_template.contacts_data",
      "output_config": "instantiated from custom_params_template.output_config"
    }
  },

  "expected_output_example": {
    "markor_files": [
      {
        "file_name": "Vendor_Contract_Sync_agenda.md",
        "content": "# Vendor Contract Sync\n- Date: 2023-10-24\n- Time: 10:00\n- Location: Meeting Room 2B\n- Attendees: Nina Patel, Omar Reyes\n\n## Related Tasks\n- Draft agenda for vendor sync\n- Collect renewal milestones\n"
      }
    ],
    "sms_sent": [
      {
        "to": "Nina Patel (+13661100111)",
        "for_meeting": "Vendor Contract Sync",
        "body": "Agenda for Vendor Contract Sync: Draft agenda for vendor sync; Collect renewal milestones. Location: Meeting Room 2B."
      },
      {
        "to": "Omar Reyes (+13661100112)",
        "for_meeting": "Vendor Contract Sync",
        "body": "Agenda for Vendor Contract Sync: Draft agenda for vendor sync; Collect renewal milestones. Location: Meeting Room 2B."
      }
    ]
  }
}
\end{lstlisting}

\clearpage
{\small
\bibliographystyle{plainnat}
\bibliography{Bib}
}


\end{document}